\documentclass[sigconf, nonacm]{acmart}

\AtBeginDocument{%
  \providecommand\BibTeX{{%
    \normalfont B\kern-0.5em{\scshape i\kern-0.25em b}\kern-0.8em\TeX}}}

\RequirePackage{algorithm}
\RequirePackage{algorithmic}
\usepackage{enumitem}
\usepackage{adjustbox} 
\usepackage{todonotes}

\begin{document}

\title{Chasing Your Long Tails: Differentially Private Prediction in Health Care Settings}

\author{Vinith M. Suriyakumar, Nicolas Papernot, Anna Goldenberg, Marzyeh Ghassemi}
\email{vinith@cs.toronto.edu}
\affiliation{%
  \institution{University of Toronto, Vector Institute}
}


\begin{abstract}

Machine learning models in health care are often deployed in settings where it is important to protect patient privacy.
In such settings, methods for \emph{differentially private} (DP) learning provide a general-purpose approach to learn models with privacy guarantees. Modern methods for DP learning ensure privacy through mechanisms that censor information judged as too unique. The resulting privacy-preserving models therefore neglect information from the \emph{tails} of a data distribution, resulting in a loss of accuracy that can disproportionately affect small groups. In this paper, we study the effects of DP learning in health care. We use state-of-the-art methods for DP learning to train privacy-preserving models in clinical prediction tasks, including x-ray classification of images and mortality prediction in time series data. We use these models to perform a comprehensive empirical investigation of the tradeoffs between privacy, utility, robustness to dataset shift and fairness. Our results highlight lesser-known limitations of methods for DP learning in health care, models that exhibit steep tradeoffs between privacy and utility, and models whose predictions are disproportionately influenced by large demographic groups in the training data. We discuss the costs and benefits of differential private learning in health care.
\end{abstract}



\maketitle

\section{Introduction}
The potential for machine learning to learn clinically relevant patterns in health care has been demonstrated across a wide variety of tasks~\cite{Tomasev2019-ug,Gulshan2016-tv,wu_modeling_2019,rajkomar2018scalable}. However, machine learning models are susceptible to privacy attacks~\cite{shokri2017membership, fredrikson2015model} that allow malicious entities with access to these models to recover sensitive information, e.g., HIV status or zip code, of patients who were included in the training data. Others have shown that anonymized electronic health records (EHR) can be re-identified using simple ``linkages'' with public data~\cite{Sweeney2015-po}, and that neural models trained on EHR are susceptible to membership inference attacks~\cite{shokri2017membership, jordon2020hide}.

\emph{Differential privacy} (DP) has been proposed as a leading technique to minimize re-identification risk through linkage attacks~\cite{Narayanan2008-fx, dwork2017exposed}, and is being used to collect personal data by the 2020 US Census~\cite{Hawes2020Implementing}, user statistics in iOS and MacOS by Apple~\cite{tang2017privacy}, and Chrome Browser data by Google~\cite{nguyen2016collecting}. DP is an algorithm-level guarantee used in machine learning~\cite{dwork2006calibrating}, where an algorithm is said to be differentially private if its output is statistically indistinguishable when applied to two input datasets that differ by only one record in the dataset.
DP learning focuses with increasing intensity on learning the ``body'' of a targeted distribution as the desired level of privacy increases.
Techniques such as differentially private stochastic gradient descent (DP-SGD)~\cite{Abadi2016-lk} and objective perturbation~\cite{chaudhuri2011differentially, neel2019differentially} have been developed to efficiently train models with DP guarantees, but introduce a \emph{privacy-utility tradeoff}~\cite{geng2020tight}. This tradeoff has been well-characterized in computer vision~\cite{papernot2020making}, and tabular data ~\cite{shokri2017membership,jayaraman2019evaluating} but have not yet been characterized in health care datasets.
Further, DP learning has asymptotic theoretical guarantees about robustness that have been established~\cite{nissim2015generalization,jung2019new}, but \emph{privacy-robustness tradeoffs} have not been evaluated in health care settings. Finally, more ``unique'' minority data may not be well-characterized by DP, leading to a noted \emph{privacy-fairness tradeoff} in vision~\cite{bagdasaryan2019differential,farrand2020neither} and natural language settings~\cite{bagdasaryan2019differential}.

To date there has not been a robust characterization of utility, privacy, robustness, and fairness tradeoffs for DP models in health care settings. Patient health and care are often highly individualized with a heavy ``tail'' due to the complexity of illness and treatment variation~\citep{hripcsak2016characterizing}, and any loss of model utility in a deployed model is likely to hinder delivered care~\cite{topol2019high}. 
Privacy-robustness tradeoffs may also be high cost in health care, as data is highly volatile and variant, evolving quickly in response to new conditions~\cite{cohen2020covid}, clinical practice shifts~\cite{herrera2019meta}, and underlying EHR systems changing~\cite{Nestor2019-zi}. 
Privacy-fairness tradeoffs are perhaps the most pernicious concern in health care as there are well-documented prejudices in health care~\cite{chen2020ethical}. Importantly, the data of patients from minority groups also often lie even further in data tails because lack of access to care can impact patients' EHR presence~\cite{ferryman2018fairness}, and leads to small sample sizes of non-white patients~\cite{chen2018my}. 

In this work, we investigate the feasibility of using DP methods to train models for health care tasks. We characterize the impact of DP in both linear and neural models on 1) accuracy, 2) robustness, and 3) fairness. First, we establish the privacy-utility tradeoffs within two health care datasets (NIH Chest X-Ray data~\cite{wang2017chestx}, and MIMIC-III EHR data~\cite{johnson_mimic-iii_2016}) as compared to two vision datasets (MNIST~\cite{lecun2010mnist} and Fashion-MNIST~\cite{xiao2017fashion}). 
We find that DP models have severe privacy-utility tradeoffs in the MIMIC-III EHR setting, using three common tasks --- mortality, long-length of stay (LOS), and an intervention (vasopressor) onset~\cite{harutyunyan_multitask_2017, wang_mimic-extract:_2019}. Second, we investigate the impact of DP on robustness to dataset shifts in EHR data. Because medical data often contains dataset shifts over time~\cite{ghassemi2018opportunities}, we create a realistic yearly model training scenario  
and evaluate the robustness of DP models under these shifts. Finally, we investigate the impact of DP on fairness in two ways: loss of performance and loss of influence. We define loss of performance through the standard, and often competing, group fairness metrics~\cite{hardt2016equality, kearns2018preventing, kearns2019empirical} of performance gap, parity gap, recall gap, and specificity gap. We examine fairness further by looking at loss of minority data importance with influence functions~\cite{koh2017understanding}.
In our audits, we focus on loss of population minority influence, e.g., importance of Black patient data, and label minority influence, e.g, importance of positive class patient data, across levels of privacy (low to high) and levels of dataset shift (least to most malignant).

Across our experiments we find that DP learning algorithms are \emph{not} well-suited for off-the-shelf use in health care. First, DP models have significantly more severe privacy-utility tradeoffs in the MIMIC-III EHR setting, and this tradeoff is proportional to the size of the tails in the data. This tradeoff holds even in larger datasets such as NIH Chest X-Ray. We further find that DP learning does not increase model robustness in the presence of small or large dataset shifts, despite theoretical guarantees~\cite{jung2019new}. Finally, we do not find a significant drop in standard group fairness definitions, unlike other domains~\cite{bagdasaryan2019differential}, likely due to the dominating effect of utility loss. We do, however, find a large drop in minority class influence. Specifically, we show that Black training patients lose ``helpful'' influence on Black test patients. Finally, we outline a series of open problems that future work should address to make DP learning feasible in health care settings.

\subsection{Contributions}
In this work, we evaluate the impact of DP learning on linear and neural networks across three tradeoffs: privacy-utility, privacy-robustness and privacy-fairness. 
Our analysis contributes to a call for ensuring that privacy mechanisms equally protect all individuals~\cite{ekstrand2018privacy}.
We present the following contributions:
\begin{itemize}[leftmargin=*]
    \item \textbf{Privacy-utility tradeoffs scale sharply with tail length.} We find that DP has particularly strong tradeoffs as tasks have fewer positive examples, resulting in unusable classifier performance. Further, increasing the dataset size does not improve utility tradeoffs in our health care tasks.
    \item \textbf{There is no correlation between privacy and robustness in EHR shifts.} We show that DP generally does not improve shift robustness, with the mortality task as one exception. Despite this, we find no correlation between increasing privacy and improved shift robustness in our tasks, most likely due to the poor utility tradeoffs. 
    \item \textbf{DP gives unfair influence to majority groups that is hard to detect with standard measures of group fairness.} We show that increasing privacy does not result in disparate impact for minority groups across multiple protected attributes and standard group fairness definitions because the privacy-utility tradeoff is so extreme. We use influence functions to demonstrate that the inherent group privacy property of DP results in large losses of influence for minority groups across patient class label, and patient ethnicity labels. 
\end{itemize}

\begin{table*}[]
\begin{center}
\begin{small}
\begin{sc}
\begin{adjustbox}{max width=\textwidth}
    \centering
    \begin{tabular}{lcccccccc}
    \toprule
      Dataset & Data Type & Outcome Variable & $n$ & $d$ & Classification Task  & Tail Size & Protected Attributes & Evaluation \\
      \midrule
      \textbf{health care} \\
      \midrule
      \texttt{mimic\_mortality} & Time Series & in-ICU mortality & 21,877 & (24,69) & Binary & Large & Ethnicity & U,R, F\\ 
      \midrule
      \texttt{mimic\_los\_3} & Time Series & length of stay > 3 days & 21,877 & (24,69) & Binary & Small & Ethnicity & U,R, F \\
      \midrule
      \texttt{mimic\_intervention} & Time Series & vasopressor administration & 21,877 & (24,69) & Multiclass (4) & Small & Ethnicity & U,R, F \\
      \midrule
      \texttt{NIH\_chest\_x\_ray} & Imaging & multilabel disease prediction & 112,120 & (256,256) & Multiclass multilabel (14) & Largest & Sex & U,F\\
      \midrule
      \textbf{Vision Baselines} \\
      \midrule
      \texttt{mnist} & Imaging & number classification & 60,000 & (28,28) & Multiclass (10) & None & N/A & U \\ 
      \midrule
      \texttt{fashion\_mnist} & Imaging & clothing classification & 60,000 & (28,28) & Multiclass (10) & None & N/A & U \\ 
      \bottomrule
    \end{tabular}
\end{adjustbox}
\end{sc}
\end{small}
\caption{We analyze tradeoffs in two vision baseline datasets and two health care datasets. We use three prediction tasks in MIMIC-III with different tail sizes and focus our utility (U), robustness (R), and fairness (F) analyses on these tasks. Finally, we choose NIH Chest X-Ray which is a larger dataset with the largest tail to examine whether increasing the dataset size has an impact on utility and fairness tradeoffs.}
\label{tab:datasets}
\end{center}
\end{table*}
\section{Related Work}
\subsection{Differential Privacy} 
DP provides much stronger privacy guarantees over methods such as k-anonymity~\cite{sweeney2002k} and t-closeness~\cite{li2007t}, to a number of privacy attacks such as reconstruction, tracing, linkage, and differential attacks~\cite{dwork2017exposed}. The outputs of DP analyses are resistant to attacks based on auxiliary information, meaning they cannot be made less private~\cite{dwork2014algorithmic}. Such benefits have made DP a leading method for ensuring privacy in consumer data settings~\cite{Hawes2020Implementing,tang2017privacy,nguyen2016collecting}.
Further, theoretical analyses have demonstrated improved generalization guarantees for out of distribution examples~\cite{jung2019new}, but there has been no empirical analysis of DP model robustness, e.g., in the presence of dataset shift. 
Other theoretical analyses demonstrate that a model that is both private and approximately fair can exist in finite sample access settings. However, they show that it is impossible to achieve DP and exact fairness with non-trivial accuracy~\cite{cummings2019compatibility}. This is empirically shown in DP-SGD which has disparate impact on complex minority groups in vision and NLP~\cite{bagdasaryan2019differential, farrand2020neither}.

\paragraph{Differential Privacy in Health Care} 
Prior work on DP in machine learning for health care has focused on the distributed setting, where multiple hospitals collaborate to learn a model~\cite{beaulieu2018privacy, pfohl2019federated}. This work has shown that DP learning leads to a loss in model performance defined by area under the receiver operator characteristic (AUROC). We instead focus on analyzing the tradeoffs between privacy, robustness, and fairness, with an emphasis on the impact that DP has on subgroups. 

\subsection{Utility, Robustness, and Fairness in Health Care} 
\paragraph{Utility Needs in Health Care Tasks} 
Machine learning in health care is intended to support clinicians in their decision making, which suggests that models need to perform similarly to physicians~\cite{davenport2019potential}. The specific metric is dependent on the the task as high positive predictive value may be preferred over high negative predictive value~\cite{kelly2019key}. In this work, we focus on predictive accuracy as AUROC and AUPRC, characterizing this loss as privacy levels increase. 

\paragraph{Robustness to Dataset Shift} 
The effect of dataset shift has been studied in non-DP health care settings, demonstrating that model performance often deteriorates when the data distribution is non-stationary~\cite{jung2015implications, davis2017calibration,subbaswamy2018preventing}. Recent work has demonstrated that performance deteriorates rapidly on patient LOS and mortality prediction tasks in the MIMIC-III EHR dataset, when trained on past years, and applied to a future year~\cite{Nestor2019-zi}. We focus on this setting for a majority of our experiments, leveraging year-to-year changes in population as small dataset shifts, and a change in EHR software between 2008 and 2009 as a large dataset shift.

\paragraph{Group Fairness} 
Disparities exist between white and Black patients, resulting in health inequity in the U.S.A~\cite{orsi2010Black, obermeyer2019dissecting}. Further, even the use of some sensitive data like ethnicity in medical practice is contentious~\cite{vyas2020hidden}, and has been called into question in risk scores, for instance in estimating kidney function~\cite{martin2011color,eneanya2019reconsidering}. 

Much work has described the ability of machine learning models to exacerbate disparities between protected groups~\cite{chen2018my}; even state-of-the-art chest X-Ray classifiers demonstrate diagnostic disparities between sex, ethnicity, and insurance type~\cite{seyyed2020chexclusion}. We leverage recent work in measuring the group fairness of machine learning models for different statistical definitions~\cite{hardt2016equality} in supervised learning. 

We complement these standard metrics by also examining loss of data importance through influence functions~\cite{koh2017understanding}; influence functions have also been extended to approximate the effects of subgroups on a model's prediction~\cite{koh2019accuracy}. They demonstrate that memorization is required for small generalization error on long tailed distributions~\cite{feldman2020does}.

\section{Data}
Details of each data source and prediction task are shown in Table~\ref{tab:datasets}. The four datasets are intentionally of different sizes, with respective tasks that represent distributions with and without long tails.

\subsection{Vision Baselines} 
We use MNIST \cite{lecun2010mnist} and FashionMNIST \cite{xiao2017fashion} to demonstrate the benchmark privacy-utility tradeoffs in non-health settings with no tails. 
We use the NIH Chest X-Ray dataset~\cite{wang2017chestx} (112,120 images, details in Appendix~\ref{appendix:NIH_preproc}) to benchmark privacy-utility tradeoffs in a medically based, but still vision-focused, setting with the largest tails of all of our tasks. 

\subsection{MIMIC-III Time Series EHR Data} 
For the remainder of our analyses on privacy-robustness and privacy-fairness, we use the MIMIC-III database~\cite{johnson_mimic-iii_2016}---a publicly available anonymized EHR dataset of intensive care unit (ICU) patients (21,877 unique patient stays, details in Appendix~\ref{appendix:mimic_preproc}). We focus on two binary prediction tasks of predicting \textbf{(1)} ICU mortality (class imbalanced), \textbf{(2)} LOS greater than 3 days (class balanced) and choose one multiclass prediction tasks of predicting intervention onset for \textbf{(3)} vasopressor administration (class balanced)~\cite{harutyunyan_multitask_2017,wang_mimic-extract:_2019}. 

\paragraph{Source of Distribution Shift}
In MIMIC-III, there is a known source of dataset shift after 2008 due to a transition in the EHR used~\cite{mimicweb}. There are also smaller shifts in non-transition years as the patient distribution is non-stationary~\cite{Nestor2019-zi}. 

\section{Methodology}
We use both DP-SGD and objective perturbation across three different privacy levels to evaluate the impact that DP learning has on utility and robustness to dataset shift. Given the worse utility and robustness tradeoffs using objective perturbation, we focus our subsequent fairness analyses on DP-SGD in health care settings.  

\subsection{Model Classes} 
\paragraph{Vision Baselines} We use different convolutional neural network architectures for the MNIST and FashionMNIST prediction tasks based on prior work~\cite{papernot2020making}.
We use DenseNet-121 pretrained on ImageNet for the NIH Chest X-Ray experiments~\cite{seyyed2020chexclusion}. 

\paragraph{MIMIC EHR Tasks} For the MIMIC-III health care tasks analyses,
we choose one linear model and one neural network per task, based on the best baselines, trained without privacy, outlined in prior work creating benchmarks for the MIMIC-III dataset~\cite{wang_mimic-extract:_2019}.
For binary prediction tasks we use logistic regression (LR) \cite{cox1972regression} and gated recurrent unit with decay (GRU-D)~\cite{che_recurrent_2018}. 
For our multiclass prediction task, we use LR and 1D convolutional neural networks.

\begin{table*}[htb!]
\begin{center}
\begin{large}
\begin{sc}
\begin{adjustbox}{max width=\textwidth}
\begin{tabular}{llccc}
\toprule
\textbf{Vision Baselines} \\
\midrule
Dataset & Model & None ($\epsilon,\delta$) & Low ($\epsilon,\delta$) & High ($\epsilon,\delta$)\\
\midrule
MNIST & CNN & $98.83 \pm 0.06~(\infty, 0)$ & $98.58 \pm 0.06~(2.6\cdot{10^5})$ & $93.78 \pm 0.25~(2.01)$ \\
FashionMNIST & CNN & $87.92 \pm 0.19~(\infty, 0)$ & $87.90 \pm 0.16~(2.6\cdot{10^5})$ & $79.53 \pm 0.10~(2.01)$ \\
\midrule
\textbf{MIMIC-III} \\
\midrule
Task & Model & None ($\epsilon,\delta$) & Low ($\epsilon,\delta$) & High ($\epsilon,\delta$)\\
\midrule
Mortality & LR & $0.82 \pm 0.03~(\infty, 0)$ & $0.76 \pm 0.05~(3.50\cdot{10^5}, 10^{-5})$ & $0.60 \pm 0.04~(3.54, 10^{-5})$ \\
& GRUD & $0.79 \pm 0.03~(\infty, 0)$ & $0.59 \pm 0.09~(1.59\cdot{10^5}, 10^{-5})$ & $0.53 \pm 0.03~(2.65, 10^{-5})$ \\
\midrule
Length of Stay > 3 & LR & $0.69 \pm 0.02~(\infty, 0)$ & $0.66 \pm 0.03~(3.50\cdot{10^5}, 10^{-5})$ & $0.60 \pm 0.04~(3.54, 10^{-5})$ \\
& GRUD & $0.67 \pm 0.03~(\infty, 0)$ & $0.63 \pm 0.02~(1.59\cdot{10^5}, 10^{-5})$ & $0.61 \pm 0.03~(2.65, 10^{-5})$ \\
\midrule
Intervention Onset (Vaso) & LR & $0.90 \pm 0.03~(\infty, 0)$ & $0.87 \pm 0.03~(1.63\cdot{10^7}, 10^{-5})$ & $0.77 \pm 0.05~(0.94, 10^{-5})$ \\
& CNN & $0.88 \pm 0.04~(\infty, 0)$ & $0.86 \pm 0.02~(5.95\cdot{10^7}, 10^{-5})$ & $0.68 \pm 0.04~(0.66, 10^{-5})$ \\
\midrule
\textbf{NIH Chest X-Ray} \\
\midrule
Metric & Model & None ($\epsilon,\delta$) & Low ($\epsilon,\delta$) & High ($\epsilon,\delta$) \\
\midrule
Average AUC & DenseNet-121 & $0.84 \pm 0.00~(\infty, 0)$ & $0.51 \pm 0.01~(1.74\cdot{10^5})$ & $0.49 \pm 0.00~(0.84)$ \\
Best AUC & DenseNet-121 & $0.98 \pm 0.00$ (Hernia) & $0.54 \pm 0.04$ (Edema) & $0.54 \pm 0.05$ (Pleural Thickening) \\
Worst AUC & DenseNet-121 & $0.72 \pm 0.00$ (Infiltration) & $0.48 \pm 0.02$ (Fibrosis) & $0.47 \pm 0.02$ (Pleural Thickening) \\
\bottomrule
\end{tabular}
\end{adjustbox}
\end{sc}
\end{large}
\caption{Health care tasks have a significant tradeoff between the High and Low or None setting. The tradeoff is better in tasks with small tails (length of stay and intervention onset), and worst in tasks such as mortality and NIH Chest X-Ray with long tails. We provide the $\epsilon,\delta$ guarantees in parentheses, where $\epsilon$ represents the privacy loss (lower is better) and $\delta$ represents the probability that the guarantee does not hold (lower is better).}
\label{tab:mimic_accur}
\end{center}
\end{table*}
\subsection{Differentially Private Training} 
We train models without privacy guarantees using stochastic gradient descent (SGD). DP models are trained with  DP-SGD~\cite{Abadi2016-lk}, which is the de-facto approach for both linear models and neural networks. 
We choose not to train models using PATE~\cite{Papernot2016-oj}, because it requires access to public data for semi-supervised learning and this is unrealistic in health care settings. 
In the Appendix, we provide results for models trained using objective perturbation~\cite{chaudhuri2011differentially, neel2019differentially} which provides $(\epsilon,0)$-DP. It is only applicable to our linear models.
We focus on DP-SGD due to its more optimal theoretical guarantees \cite{jagielski2020auditing} regarding privacy-utility tradeoffs, and objective perturbation's limited applicability to linear models.
The modifications made to SGD involve clipping gradients computed on a per-example basis to have a maximum $\ell_2$ norm, and then adding Gaussian noise to these gradients before applying parameter updates~\cite{Abadi2016-lk} (Appendix~\ref{appendix:DP-SGD}).

We choose three different levels of privacy to measure the effect of increasing levels of privacy by varying levels of epsilon. We selected these levels based on combinations of the noise level, clipping norm, number of samples, and number of epochs. Our three privacy levels are: None, Low (Clip Norm = 5.0, Noise Multiplier = 0.1), and High (Clip Norm = 1.0, Noise Multiplier = 1.0). We provide a detailed description of training setup in terms of hyperparameters and infrastructure in Appendix~\ref{appendix:training_setup}.

\subsection{Privacy Metrics} We measure DP using the $\epsilon$ bound derived analytically using the Renyi DP accountant for DP-SGD. Larger values of $\epsilon$ reflect lower privacy. Note that the privacy guarantees reported for each model are underestimates because they do not include the privacy loss due to hyperparameter seearches~\cite{chaudhuri2013stability, liu2019private}.

\section{Privacy-Utility Tradeoffs}
\label{sec:utility}
We analyze the privacy-utility tradeoff by training linear and neural models with DP learning. We analyze performance across three privacy levels for the vision, MIMIC-III and NIH Chest X-Ray datasets. The privacy-utility tradeoffs for these datasets and tasks have \textit{not} been characterized yet. Our work provides a benchmark for future work on evaluating DP learning. 

\paragraph{\bf Experimental Setup}
We train both linear and neural models on the tabular MIMIC-III tasks. We train deep neural networks on NIH Chest X-Ray image tasks and the vision baseline tasks.
We first analyze the effect that increased tail length in MIMIC-III has on the privacy-utility tradeoff. Next, we compare whether linear or neural models have better privacy-utility tradeoffs. Finally, we use the NIH Chest X-Ray dataset to evaluate if increasing dataset size, while keeping similar tail sizes, results in better tradeoffs.  

\paragraph{\bf Time Series Utility Metrics} For MIMIC-III, we average the model AUROC across all shifted test sets to quantitatively measure the utility tradeoff. We measure the privacy-utility tradeoff based on the difference in performance metrics as the level of privacy increases. The average performance across years is used because it incorporates the performance variability between each of the years due to dataset shift. Results for AUPRC for MIMIC-III can be found in Appendix~\ref{appendix:mimic_auprc}. Both our AUROC and AUPRC results show extreme utility tradeoffs in health care tasks. Both metrics are commonly used to evaluate clinical performance of diagnostic tests~\cite{hajian2013receiver}. 

\paragraph{\bf Imaging Utility Metrics} For the NIH Chest X-Ray experiments, the task we experiment on is multiclass multilabel disease prediction. We average the AUROC across all 14 disease labels. For the MNIST and FashionMNIST vision baselines, the task we experiment on is multiclass prediction (10 labels for both) where we evaluate using accuracy. 

\begin{figure*}[htb]
    \centering
    \includegraphics[width=\textwidth]{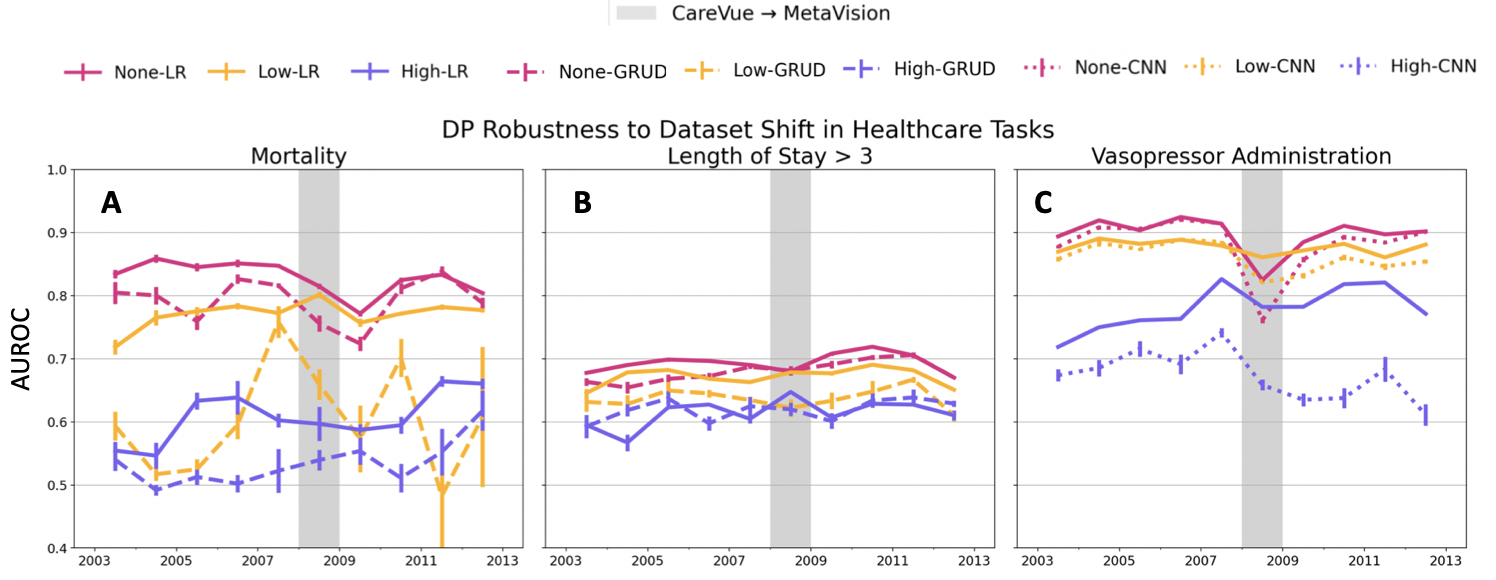}
    \caption{The effect of DP learning on robustness to non-stationarity and dataset shift. One instance of increased robustness in the 2009 column for mortality prediction in the high privacy setting (A), but this does not hold across all tasks and models. Performance drops in the 2009 column for LOS in both LR and GRU-D (B), and a much worse drop in the high privacy CNN for intervention prediction (C).}
    \label{fig:mimic_priv_robust}
\end{figure*}

\subsection{Health Care Tasks Have Steep Utility Tradeoffs}
We compare the privacy-utility tradeoffs in
Table~\ref{tab:mimic_accur}. DP-SGD generally has a negative impact on model utility. The extreme tradeoffs in MIMIC-III mortality prediction, and NIH Chest X-Ray diagnosis exemplify the information DP-SGD looses from the tails, because the positive cases are in the long tails of the distribution. There is a 22\% and 26\% drop in the AUROC between no privacy and high privacy settings for mortality prediction for LR and GRUD respectively. There is a 35\% drop in AUROC between the no privacy and high privacy settings for the NIH Chest X-Ray prediction task which has a much longer tail than mortality prediction. Our results for objective perturbation show worse utility tradeoffs than those presented by DP-SGD (Appendix~\ref{appendix:obj_pert_util}).

\subsection{Linear Models Have Better Privacy-Utility Tradeoffs} 
Across all three prediction tasks in the MIMIC-III dataset we find that linear models have better tradeoffs in the presence of long tails. This is likely due to two issues: small generalization error in neural networks often requires memorization in long tail settings~\cite{zhang2016understanding,feldman2020does} and gradient clipping introduces more bias as the number of model parameters increases~\cite{chen2020understanding,song2020characterizing}.

\subsection{Larger Datasets Do Not Achieve Better Tradeoffs}
Theoretical analyses show that privacy-utility tradeoff can be improved with larger datasets~\cite{vadhan2017complexity}. We find that the NIH Chest X-Ray dataset also has extreme tradeoffs. Despite its larger size, the dataset's positive labels in long tails are similarly lost. 

\section{Privacy-Robustness Tradeoffs}
\label{sec:robustness_main}
A potential motivation for using DP despite extreme utility tradeoffs are the recent theoretical robustness guarantees~\cite{jung2019new}.
We investigate the impact of DP to mitigating dataset shift for time series MIMIC-III tasks by analyzing model performance across years of care. 
We first record generalization as the difference in performance when a model is trained and tested on data drawn from $p$, versus performance on a shifted test set drawn from $q$) and the malignancy of the shift.
We then measure the malignancy of the yearly shifts using a domain classifier. Finally we perform a Pearsons correlation test~\cite{stigler1989francis} between the model's generalization capacity and the shift malignancy. 

\paragraph{\bf Experimental Setup}
We analyze the robustness of DP models to dataset shift in the MIMIC-III health care tasks. We use year-to-year variation in hospital practices as a small shifts, and a change in EHR software between 2008-2009 as a source of major dataset shift. We define robustness as the difference in test accuracy between in-distribution and out-distribution data. For instance, to measure model robustness from the 2006 to 2007, we would 1) \emph{train} a model on data from 2006, 2) \emph{test} the model on data from 2006, and 3) test the same model on data from 2007. The difference in these two test accuracies is the 2006-2007 model robustness. 
\begin{figure*}[htb!]
    \centering
    \includegraphics[width=\textwidth]{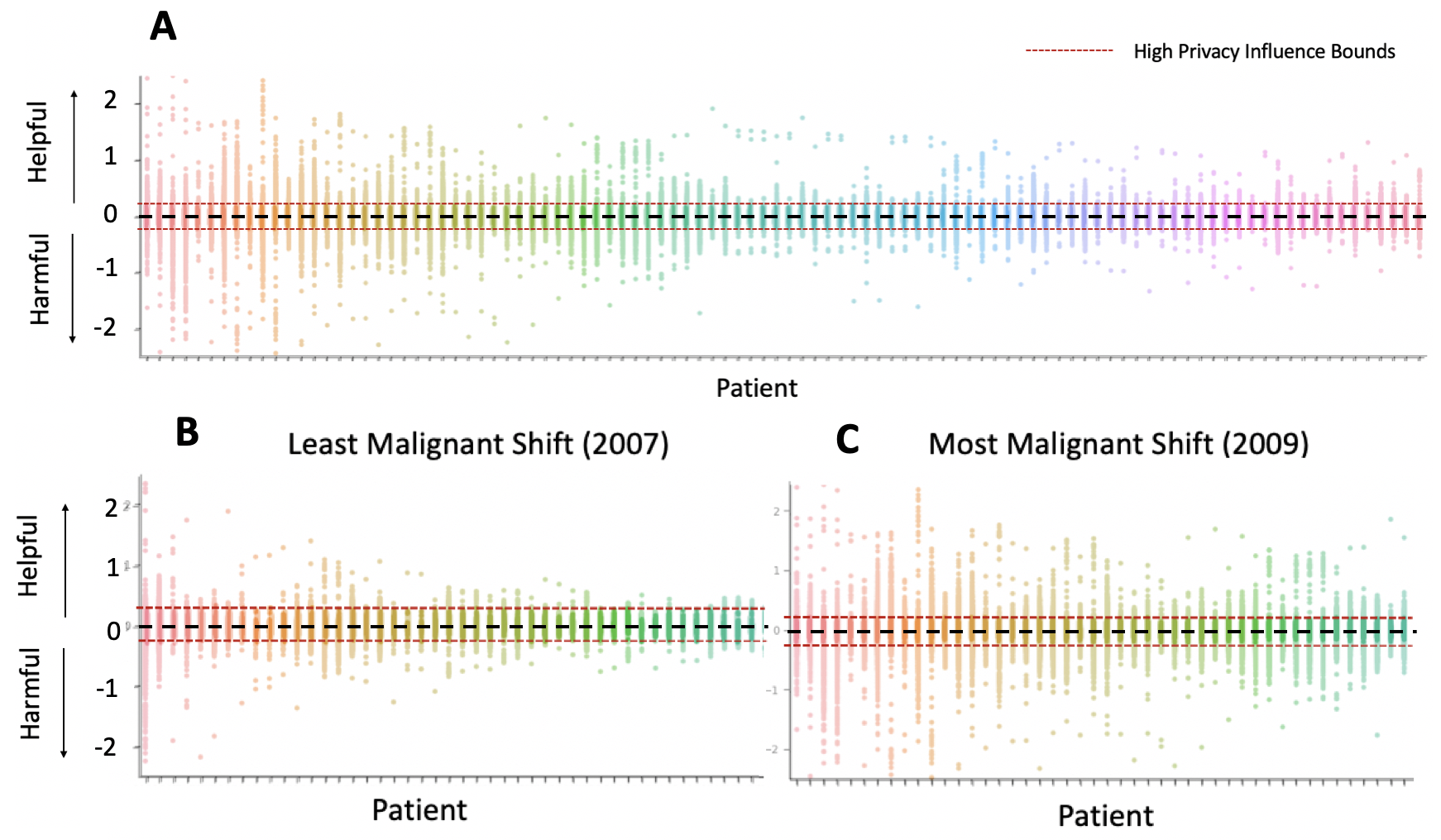}
    \caption{DP bounds the individual influence of training patients on the loss of test patients (A) which improved robustness for mortality prediction between the least malignant shift in 2007 (B) and the most malignant in 2009 (C). Individual influence of training data in the no privacy setting on 100 test patients with highest influence variance. Each column on the x-axis is an individual test patient. A unique colour is plotted per column/test patient for ease of assessment. The influence value of each patient in the training set on a specific test point is plotted as a point in that patient's column. Influence of training points is bounded in the high privacy setting (red dotted line).}
    \label{fig:no_vs_high_inf}
\end{figure*}
\vspace{-0.2cm}
\paragraph{\bf Robustness Metrics}
To measure the impact of DP-SGD on robustness to dataset shift, we measure the malignancy of yearly shifts from 2002 to 2012 for the MIMIC-III dataset. 
We then measure the correlation between malignancy of yearly shift and model performance. 
As done by others we use We use a binary domain classifier (model class is chosen best on data type) trained to discriminate between in-domain $p$ and out-domain $q$. The malignancy of the dataset shift is proportional to how difficult it is to train on $p$ and perform well in $q$~\cite{rabanser2019failing}.
Other methods such as multiple univariate hypothesis testing or multivariate hypothesis testing assume that the data is i.i.d~\cite{rabanser2019failing}. 
A full procedure is given in Appendix A, with complete significance and malignancies for each year in Appendix~\ref{appendix:shift}.

\subsection{DP Does Not Impart Robustness to Yearly  EHR Data Shift}
While we expect that DP will be more robust to dataset shift across all tasks and models, we find that model performance drops when the EHR shift occurs (2008-2009) across all privacy levels and tasks (Fig.~\ref{fig:mimic_priv_robust}). We note one exception: high privacy models are more stable in the mortality task during more malignant shifts (2007-2009) (Fig.~\ref{fig:mimic_priv_robust}).\footnote{We did not observe this improvement when training with objective perturbation  (Appendix~\ref{appendix:obj_pert_robust}).} Despite this, we find that there are no significant correlations between model robustness and privacy level (Table~\ref{tab:mimic_pearson}).  

Our analyses find that the robustness guarantees that DP provides do not hold in a large, tabular EHR setting. We note that the privacy-utility tradeoff from Section \ref{sec:utility} is too extreme in health care to conclusively understand the effect on model robustness.

\section{Privacy-Fairness Tradeoffs}
\label{sec:fairness}
Prior work has demonstrated that DP learning has disparate impact on complex minority groups in vision and NLP~\cite{bagdasaryan2019differential}. We expect similar disparate impacts on patient minority groups in the MIMIC-III and NIH Chest X-Ray datasets, based on known disparities in treatment and health care delivery~\cite{orsi2010Black, obermeyer2019dissecting}. We evaluate disparities based on four standard group fairness definitions, and on loss of minority patient influence.

We focus on the disparities between white and Black patients in MIMIC-III, based on prior work showing classifier variation in the low number of Black patients~\cite{chen2018my}.
We focus on male and female patients in NIH Chest X-Ray based on prior work exposing disparities in chest x-ray classifier performance between these two groups~\cite{seyyed2020chexclusion}. 

\begin{table*}[htb!]
\begin{center}
\begin{large}
\begin{sc}
\begin{adjustbox}{max width=\textwidth}
\begin{tabular}{lcccc}
\toprule
Privacy Level & Average Survived Influence & Average Died Influence & Most Helpful Group & Most Harmful Group Influence \\
\midrule
None & $-1.07 \pm 7.25$ & $2.28 \pm 6.91$ & \textbf{Died} & Survived\\
\midrule
Low & $-0.34 \pm 0.95$ & $0.03 \pm 0.18$ & Survived & Survived\\
\midrule
High & $-0.14 \pm 4.69$ & $0.04 \pm 1.34$ & \textbf{Survived} & Survived\\
\bottomrule
\end{tabular}
\end{adjustbox}
\end{sc}
\end{large}
\caption{Group influence summary statistics of training data by class label in all privacy levels for all test patients. Privacy changes the most helpful group the patients who died (minority) to the patients who survived (majority). DP learning minimizes the helpful influence of minority groups resulting in worse utility.}
\label{tab:utility_inf_table}
\end{center}
\end{table*}
\begin{figure*}[h]
    \centering
    \includegraphics[width=\textwidth]{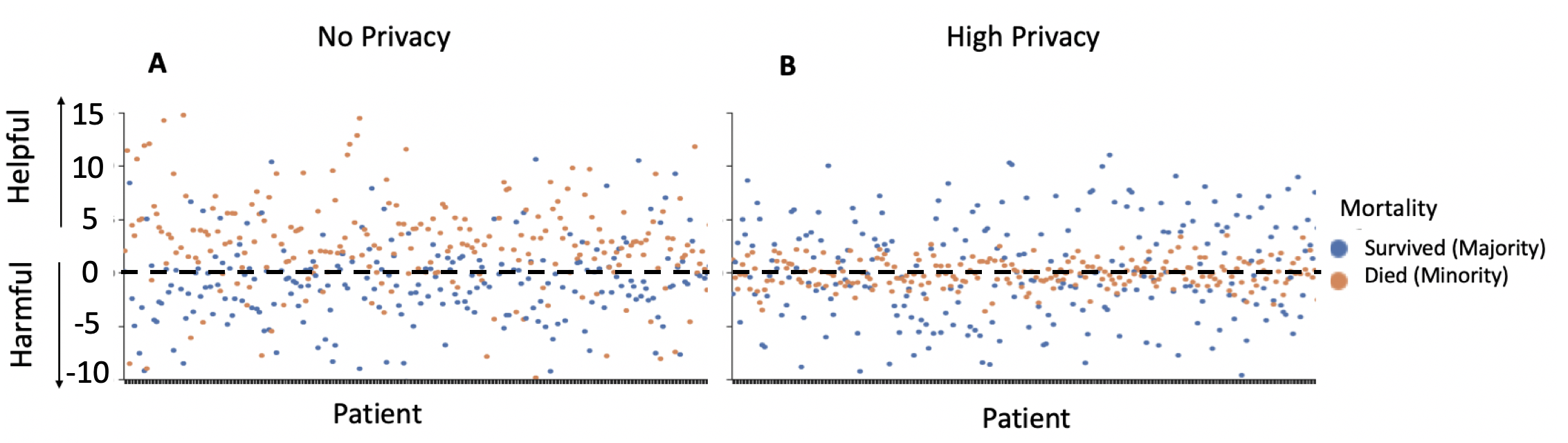}
    \caption{Group influence of training data by class label in no privacy (A) and high privacy (B) settings on 100 test patients with highest influence variance. In the no privacy setting, patients who died have a helpful influence despite being a minority class. High privacy gives the majority group the most influence due to the group privacy guarantee.}
    \label{fig:utility_inf}
\end{figure*}
\vspace{-0.3cm}

\begin{table*}[htb!]
\begin{center}
\begin{large}
\begin{sc}
\begin{adjustbox}{max width=\textwidth}
\begin{tabular}{lcccc}
\toprule
White Test Patients \\
\midrule
Privacy Level & Average White Influence & Average Black Influence & Most Helpful Ethnicity & Most Harmful Ethnicity \\
\midrule
None & $0.29 \pm 2.40$ & $0.71 \pm 1.40$ & White & White \\
\midrule
Low & $-0.22 \pm 0.70$ & $-0.03 \pm 0.17$ & White & White \\
\midrule
High & $-0.11 \pm 3.94$ & $0.03 \pm 1.35$ & White & White \\
\midrule
Black Test Patients \\
\midrule
Privacy Level & Average White Influence & Average Black Influence & Most Helpful Ethnicity & Most Harmful Ethnicity \\
\midrule
None & $0.48 \pm 1.39$ & $0.44 \pm 2.19$ & \textbf{Black} & White \\
\midrule
Low & $-0.23 \pm 0.75$ & $-0.03 \pm 0.18$ & White & White \\
\midrule
High & $-0.40 \pm 4.10$ & $0.12 \pm 1.45$ & \textbf{White} & White \\
\bottomrule
\end{tabular}
\end{adjustbox}
\end{sc}
\end{large}
\caption{Group influence summary statistics across all privacy levels for white (majority) and Black (minority) training patients on both white and Black test patients in MIMIC-III. Privacy changes the most helpful group from Black patients to the majority white patients and minimizes their helpful influence. This needs careful consideration as the use of ethnicity is still being investigated in medical practice.}
\label{tab:fairness_inf_table}
\end{center}
\vspace{-0.5cm}
\end{table*}
\paragraph{\bf Group Fairness Experimental Setup and Metrics} We measure fairness according to four standard group fairness definitions: performance gap, parity gap, recall gap, and specificity gap~\cite{hardt2016equality}. The performance gap for our health care tasks is the difference in AUROC between the selected subgroups. The remaining three definitions of fairness for binary prediction tasks are presented in Appendix~\ref{appendix:fair_def}. 

\vspace{-0.3cm}
\paragraph{\bf Influence Experimental Setup and Metrics} We use influence functions to measure the relative influence of training points on test set performance (equations in Appendix~\ref{appendix:inf_func}). Influences above 0 are \emph{helpful} in minimizing the test loss for the test patient in that column, and influences below 0 are \emph{harmful} in minimizing the test loss for that patient. Our influence function method \cite{koh2017understanding} assumes a smooth, convex loss function with respect to model parameters, and is therefore only valid for LR. We focus group privacy analyses on the LR model in no and high privacy settings for mortality prediction. 

First, we aim to confirm that the gradient clipping in DP-SGD bounds the influence of all training patients on the loss of all test patients. For the utility tradeoff, we measure the group influence that the training patients of each label group has on the loss of each test patient. For the robustness tradeoff, we measure the individual influence of all training patients on the loss of test patients between the least malignant and most malignant dataset shifts. Finally, we measure the group influence of training patients in each ethnicity on the white test patients and Black test patients separately.

\subsection{DP Has No Impact on Group Fairness on Average, But Reduces Variance Over Time}
To measure the average fairness gap in MIMIC-III, we average group fairness measures across all years of care. In the NIH Chest X-ray data we average across all disease labels. 

We find that DP-SGD has little impact on any tested fairness definitions in both MIMIC-III (Table~\ref{tbl:mimic_fairness}) and NIH Chest X-Ray, likely due to the high privacy-utility tradeoff. DP-SGD does confer lower variance in fairness measures on MIMIC-III tasks over time~(Appendix~\ref{appendix:mimic_fairness}).

\subsection{DP Learning Gives Unfair Influence to Majority Groups}
We find that DP-SGD reduces the influence of all training points on individual test points ( Fig~\ref{fig:no_vs_high_inf}) because gradient clipping tightly bounds influence of all training points across test points.

\paragraph{\bf Influence-Utility Tradeoff} We find the worst privacy-utility tradeoff in the mortality task. Non-DP models find the patients who died to be the most helpful in predictions of mortality (Fig.~\ref{fig:utility_inf} and Table~\ref{tab:utility_inf_table}). However, because positive labels, i.e., death, are rare, DP models focus influence on patients who survived, resulting in unfair over-influence. 

\paragraph{\bf Influence-Robustness Tradeoff}
We see improved robustness in LR for the mortality task which has the most malignant dataset shift (2008-2009) ( Figure~\ref{fig:mimic_priv_robust}).

We find that the variance of the influence is fairly low for non-DP models during lower malignancy shifts. In more malignant shifts, the variance of the influence is high with many training points being harmful (Fig.~\ref{fig:no_vs_high_inf}). This is likely due to gradient clipping reducing influence variance and is entangled with the poor privacy-utility tradeoff~\ref{appendix:robust_corr}.


\paragraph{\bf Influence-Fairness Tradeoff}
We approximate the collective group influence of different ethnicities in the training set on the test loss in Fig.~\ref{fig:inf_fairness_graph} and Table~\ref{tab:fairness_inf_table}. We show that group privacy results in white patients having a more significant influence, both helpful and harmful, on test patients in the high privacy setting.

\section{Discussion}

\subsection{On Utility, Robustness and Trust in Clinical Prediction Tasks}

\paragraph{\bf{Poor Utility Impacts Trust}}
While some reduced utility in long tail tasks are known~\cite{feldman2020does}, the extreme tradeoffs that we observe in Table~\ref{tab:mimic_accur} are much worse than expected. Machine learning can only support the decision making processes of clinicians if there is clinical trust. If models do not perform  well as, or better than, clinicians once we include privacy, there is little reason to trust them~\cite{topol2019high}.

\paragraph{\bf Importance of Model Robustness} Despite the promising theoretical transferability guarantees of DP, the results in Fig~\ref{fig:mimic_priv_robust} and Table~\ref{appendix:robust_corr} demonstrate these do not transfer in our health care setting. While we explored changes in EHR software as dataset shift, there are many other known shifts in healthcare data, e.g., practice modifications due to reimbursement policy changes~\cite{kocher2010affordable}, or changing clinical needs in public health emergencies such as COVID-19~\cite{cohen2020covid}. If models do not maintain their utility after dataset shifts, catastrophic silent failures could occur in deployed models~\cite{Nestor2019-zi}.  

\subsection{Majority Group Influence Is Harmful in Health Care}
We show in Figure~\ref{fig:utility_inf} that the tails of the label distribution are minority-rich, results in poor mortality prediction performance under DP. 
Prior work in evaluating model fairness in health care has focused on standard group fairness definitions~\cite{pfohl2020empirical}. However, these definitions do not provide a detailed understanding of model fairness under reduced utility. Other work has shown that large utility loss can ``wash out'' fairness impacts~\cite{farrand2020neither}. Our work demonstrates that DP learning does harm group fairness in such ``washed out'' poor utility settings by giving majority groups (e.g., those that survived, and white patients) the most influence on predictions across all subgroups. 
 
\paragraph{\bf Why Influence Matters}
Disproportionate assignment of influence is an important problem. Differences in access, practice, or recording reflect societal biases~\cite{rajkomar2018ensuring,rose2018machine}, and models trained on biased data may exhibit unfair performance in populations due to this underlying variation~\cite{chen2019can}. Further, while patients with the same diagnosis are usually more helpful for estimating prognosis in practice~\cite{croft2015science}, labels in health care often lack precision or, in some cases, may be unreliable~\cite{omalley2005measuring}. In this setting, understanding what factors are consistent high-influence in patient phenotypes is an important task~\cite{halpern2016electronic,yu2017enabling}.

\paragraph{\bf Loss Of Black Influence}
Ethnicity is currently used in medical practice as a factor in many risk scores, where different risk profiles are assumed for patients of different races \cite{martin2011color}. However, the validity of this stratification has recently been called into question by the medical community \cite{eneanya2019reconsidering}. Prior work has established the complexity of treatment variation in practice, as patient care plans are highly individualized, e.g., in a cohort of 250 million patients, 10\% of diabetes and depression patients and almost 25\% of hypertension patients had a unique treatment pathway~\cite{hripcsak2016characterizing}. Thus having the white patients become the most influential in Black patients predictions may not be desirable. 

\paragraph{\bf Anchoring Influence Loss in Systemic Injustice}
Majority over-influence is prevalent in medical settings, and has direct impact on the survival of patients. Many female and minority patients receive worse care and have worse outcomes because clinicians base their symptomatic evaluations on white and/or male patients~\cite{greenwood2018patient, greenwood2020physician}.
Further, randomized control trials (RCTs) are an important tool that 10-20\% of treatments are based on~\cite{mcginnis2013best}. However, prior work has shown that RCTs have notorious exclusive criteria for inclusion; in one salient example, only 6\% of asthmatic patients would have been eligible to enroll in the RCT that resulted in their treatments~\cite{travers2007external}. RCTs tend to be comprised of white, male patients, resulting in their data determining what is an effective treatment~\cite{heiat2002representation}. 
By removing influence from women, Hispanics, and Blacks, naive machine learning practices can exacerbate systemic injustices~\cite{chen2020ethical}.

There are ongoing efforts to improve representation of the population in RCTs, shifting away from the majority having majority influence on treatments~\cite{stronks2013confronting}. Researchers using DP should follow suit, and work to reduce the disparate impact on influence to ensure that it does not perpetuate this existing bias in health care. One solution is to start measuring individual example privacy loss~\cite{feldman2020individual} instead of a conservative worst bound across all patients. Currently, DP-SGD uses constant gradient clipping for all examples to ensure this constant worst bound. Instead, individual privacy accounting can help support adaptive gradient clipping for each example which may help to reduce the disparate impact DP-SGD has on influence. 
We also encourage future privacy-fairness tradeoff analyses to include loss of influence as a standard metric, especially where the utility tradeoff is extreme.

\subsection{Less Privacy In Tails Is Not an Option}
The straightforward solution to the long tail issue is to ``provide less or no privacy for the tails''~\cite{kearns2015privacy}. This solution could amplify existing systemic biases against minority subgroups, and minority mistrust of medical institutions. For example, Black mothers in the US are most likely to be mistreated, dying in childbirth at a rate three times higher than white women~\cite{berg1996pregnancy}. In this setting, it is not ethical choose between a ``non-private'' prediction that will potentially leak unwanted information, e.g., prior history of abortion, and a ``private'' prediction that will deliver lower quality care. 

\subsection{On the Costs and Benefits of Privacy in Health Care}

\paragraph{\bf Privacy Issues With Health Care Data} Most countries have regulations that define the protective measures to maintain patient data privacy. In North America, these laws are defined by the Health Insurance Portability and Accountability Act (HIPAA)~\cite{act1996health} in the US and Personal Information Protection and Electronic Documents Act (PIPEDA)~\cite{act2000personal} in Canada. These laws are governed by the General Data Protection Regulation (GDPR) in the EU. Recent work has shown that HIPAA's privacy regulations and standards such as anonymizing data are not sufficient to prevent advanced re-identification of data~\cite{na2018feasibility}. In one instance, researchers were able to re-identify individuals' faces from MRIs using facial recognition software~\cite{schwarz2019identification}. Privacy attacks such as these demonstrate the fear of health care data loss. 

\paragraph{\bf Who Are We Defending Against? } While there are potential concerns for data privacy, it is important realize that privacy attacks assume a powerful entity with malicious purposes~\cite{dwork2017exposed}. Patients are often not concerned when their doctors, or researchers, have access to medical data~\cite{Kalkmanmedethics-2019-105651, ghafur2020public}. However, there are concerns that private, for-profit corporations may purchases health care data that they can easily de-anonymize and link to information collected through their own products. Such linkages could result in raised insurance premiums~\cite{beaulieu2018privacy}, and unwanted targeted advertising. 
Recently, Google and University of Chicago Medicine department faced a lawsuit from a patient due to his data being shared in a research partnership between the two organizations~\cite{landi_2020}.

Setting a different standard for dataset release to for-profit entities could be one solution. This allows clinical entities and researchers to make use of full datasets without extreme tradeoffs, while addressing privacy concerns. 
\subsection{Open Problems for DP in Health Care}
While health care has been cited as an important motivation for the development of DP~\cite{papernot2020making, chaudhuri2011differentially, dwork2014algorithmic, dwork2006calibrating, wu2017bolt, vietri2020private}, our work demonstrates that it is not currently well-suited to these tasks. 
The theoretical assumptions of DP learning apply in extremely large collection settings, such as the successful deployment of DP US Census data storage. 
We highlight potential areas of research that both the DP and machine learning communities should focus on to make DP usable in health care data:
\begin{enumerate}[leftmargin=*]
    \item \textbf{Adaptive and Personalized Privacy Accounting} Many of the individuals in the body of a distribution do not end up spending as much of the privacy budget than individuals in the tails. Current DP learning methods do not account for this and simply take a constant, conservative worst case bound for everyone. Improved accounting that can give tails more influence through methods such as adaptive clipping can potentially improve the utility and fairness tradeoff. 
    \item \textbf{Auditing DP Learning in Health Care} Currently, ideal values for the $\epsilon$ guarantee are below 100 but these are often unattainable when trying to maintain utility as we demonstrate in our work. Empirically DP-SGD provides much strong guarantees against privacy attacks than those derived analytically~\cite{jagielski2020auditing}. Developing a suite of attacks for health care settings that can provide similar empirical measurement would complement analytical guarantees nicely. It would provide decision makers more realistic information about what $\epsilon$ value they actually need in health care.
\end{enumerate}

\section{Conclusion}
In this work, we investigate the feasibility of using DP-SGD to train models for health care prediction tasks. We find that DP-SGD is not well-suited to health care prediction tasks in its current formulation. First, we demonstrate that DP-SGD loses important information about minority classes (e.g., dying patients, minority ethnicities) that lie in the tails of the dat distribution. The theoretical robustness guarantees of DP-SGD do not apply to the dataset shifts we evaluated. We show that DP learning disparately impacts group fairness when looking at loss of influence for majority groups. We show this disparate impact occurs even when standard measures of group fairness show no disparate impact due to poor utility. This imposed asymmetric valuation of data by the model requires careful thought, because the appropriate use of class membership labels in medical settings in an active topic of discussion and debate. Finally, we propose open areas of research to improve the usability of DP in health care settings. Future work should target modifying DP-SGD, or creating novel DP learning algorithms, that can learn from data distribution tails effectively, without compromising privacy.  

\begin{acks}
We would like to acknowledge the following funding sources: New Frontiers in Research Fund - NFRFE-2019-00844. Resources used in preparing this research were provided, in part, by the Province of Ontario, the Government of Canada through CIFAR, and companies sponsoring the Vector Institute \hyperlink{www.vectorinstitute.ai/partners}{www.vectorinstitute.ai/partners}. Thank you to the MIT Laboratory of Computational Physiology for facilitating year of care access to the MIMIC-III database. Finally, we would like to thank Nathan Ng, Taylor Killian, Victoria Cheng, Varun Chandrasekaran, Sindhu Gowda, Laleh Seyyed-Kalantari, Berk Ustun, Shalmali Joshi, Natalie Dullerud, Shrey Jain, and Sicong (Sheldon) Huang for their helpful feedback. 
\end{acks}
\bibliographystyle{ACM-Reference-Format}
\bibliography{acmart}

\onecolumn
\section{Appendices}
\appendix

\section{Background}
\subsection{Differential Privacy}
Formally, a learning algorithm $L$ that trains models from the dataset $D$ satisfies ($\epsilon$,$\delta$)-DP if the following holds for all training datasets $d$ and $d'$ with a Hamming distance of 1:
\begin{equation}
    Pr[L(d) \in D] \leq e^{\epsilon}Pr[L(d') \in D] + \delta
\end{equation}
The parameter $\epsilon$ measures the formal privacy guarantee by defining an upper bound on the privacy loss in the worst possible case. A smaller $\epsilon$ represents stronger privacy guarantees. The $\delta$ factor allows for some probability that the property may not hold. For the privacy guarantees of a subgroup of size $k$, differential privacy defines the guarantee as:
\begin{equation}
    Pr[L(d) \in D] \leq e^{k\epsilon}Pr[L(d') \in D] + \delta
\end{equation}
This group privacy guarantee means that the privacy guarantee degrades linearly with the size of the group.
\subsection{Measuring Dataset Shift}
First, we create a balanced dataset made up of samples from both the training and test distributions. We then train a domain classifer, a binary classifier to distinguish samples from being in the training or test distribution on this dataset. We measure whether the accuracy of the domain classifier is significantly better than random chance (0.5), which means dataset shift is significant, using binomial testing. We setup the binomial test as:

\begin{equation}
    H_0: accuracy = 0.5~v.s.~ H_A: accuracy \neq 0.5
\end{equation}

Under the null hypothesis, the accuracy of the classifier follows a binomial distribution: $accuracy\sim{Bin(N_{test},0.5)}$ where $N_{test}$ is the number of samples in the test set.

Assuming that the p-value is less than 0.05 for our hypothesis test on the accuracy of our domain classifier being better than random chance, we diagnose the malignancy of the shift. We train a binary prediction classifier without privacy on the original training set of patients. Then, we select the top 100 samples that the domain classifier most confidently predicted as being in the test distribution. Finally, we determine the malignancy of the shift by evaluating the performance of our prediction classifiers on the selected top 100 samples. Low accuracy means that the shift is malignant.

\subsection{Fairness Definitions}
\label{appendix:fair_def}
\begin{table}[htb!]
\begin{sc}
\begin{adjustbox}{max width=\textwidth}
    \centering
    \begin{tabular}{l|l|l|l}
    \toprule
    Fairness Metric & Definition & Gap Name & Gap Equation \\  
    \midrule
    Demographic parity   & $P(\hat{Y} = y) = P(\hat{Y} = \hat{y}|Z=z), \forall{z} \in{Z}$  & Parity Gap & $\frac{TP_{1}+FP_{1}}{N_{1}} - \frac{TP_{2}+FP_{2}}{N_{2}}$
    \\
    \midrule
    Equality of opportunity (positive class) & $P(\hat{Y} = 1 | Y = 1) = P(\hat{Y} = 1| Y = 1, Z=z), \forall{z} \in{Z}$  & Recall Gap & $\frac{TP_{1}}{TP_{1}+FN_{1}} - \frac{TP_{2}}{TP_{2}+FN_{2}}$
    \\
    \midrule
    Equality of opportunity (negative class) & $P(\hat{Y} = 0 | Y = 0) = P(\hat{Y} = 0| Y = 0, Z=z), \forall{z} \in{Z}$  & Specificity Gap & $\frac{TN_{1}}{TN_{1}+FP_{1}} - \frac{TN_{2}}{TN_{2}+FP_{2}}$ \\
    \bottomrule
    \end{tabular}
\end{adjustbox}
\end{sc}
\caption{The three equalized odds definitions of fairness that we use in the binary prediction tasks. The recall gap is most relevant in health care where we aim to minimize false negatives.}
\label{tab:fariness_def}
\end{table}
\subsection{Influence Functions}
\label{appendix:inf_func}
Using the approach from~\cite{koh2017understanding} we analyze the influence of all training points on the loss for each test point defined in Equation~\ref{eqn:influence_1}. The approach formalizes the goal of understanding how the model's predictions would change if we removed a training point. This is a natural connection to differential privacy which confers algorithmic stability by bounding the influence that any one training point has on the output distribution. First, the approach uses influence functions to approximate the change in model parameters by computing the parameter change if $z_{train}$ was upweighted by some small $\tau$. The new parameters of the model are defined by $\hat{\theta}_{-z_{train}}$ = $argmin_{\theta{\in}{\Theta}}$ $\frac{1}{n}\Sigma_{i=1}^{n}L(x_{i},\theta) + \tau{L(z_{train},\theta)}$. Next, applying the results from \cite{cook1980characterizations} and applying the chain rule the authors achieve Equation~\ref{eqn:influence_1} to characterize the influence that a training point has the loss on another test point.  Influence functions use an additive property for interpreting the influence of subgroups showing that the group influence is the sum of the influences of all individual points in the subgroup but that this is usually an underestimate of the true influence of removing the subgroup~\cite{koh2019accuracy}.
\begin{equation}
    \label{eqn:influence_1}
    I_{up,loss}(z_{train},z_{test}) = -\nabla_{\theta}L(z_{test},\hat{\theta})^{T}H_{\hat{\theta}}^{-1}\nabla_{\theta}L(z_{train}, \hat{\theta})
\end{equation}

\section{Data Processing}
\label{appendix:preprocessing}
\subsection{MIMIC-III}
\label{appendix:mimic_preproc}
\subsubsection{Cohort}
Within the MIMIC-III dataset~\cite{johnson_mimic-iii_2016}, each individual patient may be admitted to the hospital on multiple different occasions and may be transferred to the ICU multiple times during their stay. We choose to focus on a patient's first visit to the ICU, which is the most common case. Thus, we extract a cohort of patient EHR that corresponds to first ICU visits. We also only focus on ICU stays that lasted at least 36 hours and all patients older than 15 years of age. Using the MIMIC-Extract~\cite{wang_mimic-extract:_2019} processing pipeline, this results in a cohort of 21,877 unique ICU stays. The breakdown of this cohort by year, ethnicity and class labels for each task can be found in Table~\ref{tb2:ethnicity_mort}.
\subsubsection{Features: Demographics and Hourly Labs and Vitals}
7 static demographic features and 181 lab and vital measurements which vary over time are collected for each patient's stay. The 7 demographic features comprise gender and race attributes which we observe for all patients. Meanwhile, the 181 lab results and vital signs have a high rate of missingness ($>$90.6\%) because tests are only ordered based on the medical needs of each patient. These tests also incur infrequently over time. This results in our dataset having irregularly sampled time series with high rates of missingness.
\subsubsection{Transformation to 24-hour time-series}
All time-varying measurements are aggregated into regularly-spaced hourly buckets (0-1 hr, 1-2 hr, etc.). Each recorded hourly value is the mean of any measurements captured in that hour. Each numerical feature is normalized to have zero mean and unit variance. The input to each prediction model is made up of two parts: the 7 demographic features and an hourly multivariate time-series of labs and vitals. The time series are censored to a fixed-duration of the first 24 hours to represent the first day of a patient's stay in the ICU. This means all of our prediction tasks are performed based on the first day of a patient's stay in the ICU. 
\subsubsection{Imputation of missing values}
We impute our data to deal with the high rate of missingness using a strategy called "simple imputation" developed by~\cite{che_recurrent_2018} for MIMIC time-series prediction tasks. Each separate univariate measurement is forward filled, concatenated with a binary indicator if the value was measured within that hour and concatenated with the time since the last measurement of this value.
\subsubsection{Clinical Aggregations Representation}
Data representation is known to be important in building robust machine learning models. Unfortunately, there is a lack of medical data representations that are standards compared to representations such as Gabor filters is computer vision.~\cite{Nestor2019-zi} explore four different representations for medical prediction tasks and demonstrate that their medical Aggregations representation is the most robust to temporal dataset shift. Thus, we use the medical aggregations representation to train our models. This representation groups together values that measure the same physiological quantity but are under different ItemIDs in the different EHR. This reduces the original 181 time-varying values to 68 values and reduces the rate of missingness to 78.25\% before imputation.
\subsection{NIH Chest X-Ray}
We resize all images to 256x256 and normalize via the mean and standard deviation of the ImageNet dataset. We apply center crop, random horizontal flip, and validation
set early stopping to select the optimal model. We further perform random 10 degree rotation as data augmentation. \label{appendix:NIH_preproc}

\newpage
\section{Data Statistics}
\subsection{MIMIC-III}
For understanding of the class imbalance and ethnicity frequency in the binary prediction health care tasks, a table of these statistics is provided in Table~\ref{tb2:ethnicity_mort} which show the imbalance between ethnicities across all tasks and the class imbalance in the mortality task.

\begin{table*}[h]
\begin{center}
\begin{scriptsize}
\begin{sc}
\begin{adjustbox}{max width=\textwidth}
\begin{tabular}{lcccccr}
\toprule
Ethnicity Breakdown \\
\midrule
& & Total & Mortality && Length of Stay \\
\midrule
Year & Ethnicity & & Negative & Positive & Negative & Positive \\
\midrule
2001 & Asian & 5 & 80\% & 20\% & 60\% & 40\% \\
& Black & 48 & 90\% & 10\% & 52\% & 48\% \\
& Hispanic & 7 & 86\% & 14\% & 57\% & 43\% \\
& Other & 217 & 86\% & 14\% & 53\% & 47\% \\
& White & 319 & 92\% & 8\% & 54\% & 46\% \\
\midrule
2002 & Asian & 23 & 100\% & 0\% & 48\% & 52\% \\
& Black & 102 & 92\% & 8\% & 56\% & 44\% \\
& Hispanic & 25 & 96\% & 4\% & 52\% & 48\% \\
& Other & 520 & 88\% & 12\% & 49\% & 51\% \\
& White & 937 & 93\% & 7\% & 51\% & 49\% \\
\midrule
2003 & Asian & 34 & 94\% & 6\% & 38\% & 62\% \\
& Black & 116 & 97\% & 3\% & 58\% & 24\% \\
& Hispanic & 45 & 96\% & 4\% & 53\% & 47\% \\
& Other & 465 & 90\% & 10\% & 46\% & 54\% \\
& White & 1203 & 94\% & 6\% & 50\% & 50\% \\
\midrule
2004 & Asian & 31 & 94\% & 6\% & 68\% & 32\% \\
& Black & 134 & 96\% & 4\% & 51\% & 49\% \\
& Hispanic & 38 & 89\% & 11\% & 47\% & 53\% \\
& Other & 353 & 90\% & 10\% & 45\% & 55\% \\
& White & 1236 & 93\% & 7\% & 51\% & 49\% \\
\midrule
2005 & Asian & 50 & 90\% & 10\% & 46\% & 54\% \\
& Black & 142 & 96\% & 4\% & 56\% & 44\% \\
& Hispanic & 48 & 96\% & 4\% & 56\% & 44\% \\
& Other & 279 & 89\% & 11\% & 48\% & 52\% \\
& White & 1323 & 91\% & 9\% & 51\% & 49\% \\
\midrule
2006 & Asian & 60 & 92\% & 8\% & 47\% & 53\% \\
& Black & 160 & 96\% & 4\% & 59\% & 41\% \\
& Hispanic & 62 & 97\% & 3\% & 45\% & 55\% \\
& Other & 215 & 89\% & 11\% & 49\% & 51\% \\
& White & 1434 & 93\% & 7\% & 54\% & 46\% \\
\midrule
2007 & Asian & 58 & 93\% & 7\% & 59\% & 41\% \\
& Black & 170 & 95\% & 5\% & 55\% & 45\% \\
& Hispanic & 73 & 99\% & 1\% & 59\% & 41\% \\
& Other & 235 & 88\% & 12\% & 52\% & 48\% \\
& White & 1645 & 93\% & 7\% & 54\% & 46\% \\
\midrule
2008 & Asian & 69 & 93\% & 8\% & 51\% & 49\% \\
& Black & 162 & 98\% & 2\% & 56\% & 44\% \\
& Hispanic & 90 & 93\% & 7\% & 46\% & 54\% \\
& Other & 136 & 91\% & 9\% & 56\% & 44\% \\
& White & 1691 & 93\% & 7\% & 54\% & 46\% \\
\midrule
2009 & Asian & 53 & 94\% & 6\% & 66\% & 34\% \\
& Black & 150 & 94\% & 6\% & 59\% & 41\% \\
& Hispanic & 70 & 93\% & 7\% & 59\% & 41\% \\
& Other & 180 & 87\% & 13\% & 57\% & 43\% \\
& White & 1612 & 93\% & 7\% & 55\% & 45\% \\
\midrule
2010 & Asian & 55 & 87\% & 13\% & 47\% & 53\% \\
& Black & 177 & 97\% & 3\% & 65\% & 35\% \\
& Hispanic & 71 & 96\% & 4\% & 54\% & 46\% \\
& Other & 303 & 87\% & 13\% & 52\% & 48\% \\
& White & 1568 & 94\% & 6\% & 52\% & 48\% \\
\midrule
2011 & Asian & 63 & 92\% & 8\% & 63\% & 37\% \\
& Black & 191 & 93\% & 7\% & 55\% & 45\% \\
& Hispanic & 89 & 96\% & 4\% & 53\% & 47\% \\
& Other & 268 & 85\% & 15\% & 45\% & 55\% \\
& White & 1622 & 95\% & 5\% & 54\% & 46\% \\
\midrule
2012 & Asian & 42 & 98\% & 2\% & 52\% & 48\% \\
& Black & 127 & 95\% & 5\% & 53\% & 47\% \\
& Hispanic & 55 & 93\% & 7\% & 71\% & 29\% \\
& Other & 276 & 91\% & 9\% & 54\% & 46\% \\
& White & 945 & 92\% & 8\% & 58\% & 42\% \\
\bottomrule
\end{tabular}
\end{adjustbox}
\end{sc}
\end{scriptsize}
\caption{This is a breakdown of patients in our cohort by ethnicity for each year for both tasks.}
\label{tb2:ethnicity_mort}
\end{center}
\end{table*}

\subsection{NIH Chest X-Ray}
\begin{table}[htb!]
    \centering
    \begin{tabular}{c|c|c|c|c}
    \toprule
         \# of Images & \# of Patients & View & Male & Female  \\
         112,120 & 30,805 & Front & 56.49\% & 43.51\% \\
    \bottomrule
    \end{tabular}
    \caption{Sex breakdown of images in NIH Chest X-Ray dataset}
    \label{tab:my_label}
\end{table}

\begin{table}[htb!]
    \centering
    \begin{tabular}{l|c|c}
    \toprule
         Disease Label & Negative Percentage & Positive Class Percentage  \\
         Atelectasis & 94.48\% & 5.52\% \\
         Cardiomegaly & 2.50\% & 97.50\% \\
         Consolidation & 1.40\% & 98.60\% \\
         Edema & 0.26\% & 99.74\% \\
         Effusion & 4.16\% & 85.84\% \\
         Emphysema & 0.86\% & 99.14\% \\
         Fibrosis & 1.85\% & 98.15\% \\
         Hernia & 0.27\% & 99.73\% \\
         Infiltration & 11.70\% & 88.30\% \\
         Mass & 4.16\% & 95.84\% \\
         Nodule & 5.39\% & 94.61\% \\
         Pleural Thickening & 2.48\% & 97.52\% \\
         Pneumonia & 0.55\% & 99.45\% \\
         Pneumothorax & 0.88\% & 99.22\% \\
    \bottomrule
    \end{tabular}
    \caption{Disease label breakdown of images in NIH Chest X-Ray dataset}
    \label{tab:my_label}
\end{table}

\newpage
\section{MIMIC-III: Dataset Shift Quantification}
\label{appendix:shift}
\subsection{Domain Classifier}
Using the domain classifier method presented in the main paper, we evaluate both the significance of the dataset shift and the malignancy of the shift. The shift between EHR systems is most malignant in the mortality task for LR while there are not highly malignant shifts in LOS or intervention prediction for LR (Table~\ref{tbl:lr}). Meanwhile, in GRUD the shift between the EHR systems is no longer malignant across any of the binary tasks and the shift is relatively more malignant in the intervention prediction task for CNNs (Table~\ref{tb2:nn}). The domain classifier performed significantly better than random chance if the p-value in parentheses is less than $5.0\cdot{10^{-2}}$.
\begin{table}[htb!]
\begin{center}
\begin{small}
\begin{sc}
\begin{adjustbox}{max width=\textwidth}
\begin{tabular}{lccc}
\toprule
& & Task & \\
\midrule
Year & Mortality & LOS & Intervention Prediction (Vaso) \\
\midrule
2002 & $0.67~(1.39\cdot{10^{-72}})$ & $0.64~(1.95\cdot{10^{-79}})$ & $0.93~(2.73\cdot{10^{-20}})$ \\
2003 & $0.83~(1.56\cdot{10^{-68}})$ & $0.68~(1.04\cdot{10^{-63}})$ & $0.96~(6.45\cdot{10^{-24}})$ \\
2004 & $0.86~(7.26\cdot{10^{-192}})$ & $0.59~(3.16\cdot{10^{-180}})$ & $0.89~(2.54\cdot{10^{-16}})$ \\
2005 & $0.89~(0.00)$ & $0.62~(0.00)$ & $0.94~(2.01\cdot{10^{-21}})$ \\
2006 & $0.92~(0.00)$ & $0.63~(0.00)$ & $0.95~(1.25\cdot{10^{-22}})$ \\
2007 & $0.93~(4.94\cdot{10^{-324}})$ & $0.64~(4.94\cdot{10^{-324}})$ & $0.97~(2.63\cdot{10^{-25}})$ \\
2008 & $0.72~(4.94\cdot{10^{-324}})$ & $0.65~(4.94\cdot{10^{-324}})$ & $0.91~(3.21\cdot{10^{-19}})$ \\
2009 & $0.14~(4.94\cdot{10^{-324}})$ & $0.64~(4.94\dot{10^{-324}})$ & $0.92~(3.21\cdot{10^{-19}})$ \\
2010 & $0.20~(0.00)$ & $0.54~(0.00)$ & $0.95~(1.25\cdot{10^{-22}})$ \\
2011 & $0.13~(4.94\cdot{10^{-324}})$ & $0.64~(4.94\cdot{10^{-324}})$ & $0.85~(4.83\cdot{10^{-13}})$ \\
2012 & $0.14~(4.94\cdot{10^{-324}})$ & $0.50~(4.94\cdot{10^{-324}})$ & $0.96~(6.45\cdot{10^{-24}})$ \\
\bottomrule
\end{tabular}
\end{adjustbox}
\end{sc}
\end{small}
\caption{Shift malignancy with statistical significance of domain classifier performance in parentheses. Lower values represent higher malignancy. LR was used for both the domain classifier and determining the accuracy on the top 100 most anomalous samples. The shift between EHRs is most malignant in the mortality task.}
\label{tbl:lr}
\end{center}
\end{table}
\begin{table}[htb!]
\begin{center}
\begin{small}
\begin{sc}
\begin{adjustbox}{max width=\textwidth}
\begin{tabular}{lccc}
\toprule
& & Task & \\
\midrule
Year & Mortality & LOS & Intervention Prediction (Vaso) \\
\midrule
2002 & $0.90~(1.80\cdot{10^{-186}})$ & $0.62~(1.14\cdot{10^{-165}})$ & $0.99~(1.59\cdot{10^{-28}})$ \\
2003 & $0.95~(1.28\cdot{10^{-45}})$ & $0.67~(1.97\cdot{10^{-64}})$ & $1.00~(1.58\cdot{10^{-30}})$ \\
2004 & $0.42~(3.21\cdot{10^{-290}})$ & $0.66~(6.57\cdot{10^{-317}})$ & $0.99~(1.59\cdot{10^{-28}})$ \\
2005 & $0.28~(0.00)$ & $0.55~(0.00)$ & $1.00~(1.58\cdot{10^{-30}})$ \\
2006 & $0.13~(0.00)$ & $0.56~(0.00)$ & $1.00~(1.58\cdot{10^{-30}})$ \\
2007 & $0.90~(0.00)$ & $0.49~(0.00)$ & $0.99~(1.59\cdot{10^{-28}})$ \\
2008 & $0.74~(4.94\cdot{10^{-324}})$ & $0.65~(4.94\cdot{10^{-324}})$ & $0.93~(2.73\cdot{10^{-20}})$ \\
2009 & $0.86~(4.94\cdot{10^{-324}})$ & $0.60~(4.94\cdot{10^{-324}})$ & $0.96~(6.45\cdot{10^{-24}})$ \\
2010 & $0.87~(4.94\cdot{10^{-324}})$ & $0.84~(4.94\cdot{10^{-324}})$ & $1.00~(9.60\cdot{10^{-1}})$ \\
2011 & $0.94~(0.00)$ & $0.83~(0.00)$ & $0.99~(9.60\cdot{10^{-1}})$ \\
2012 & $0.96~(0.00)$ & $0.89~(0.00)$ & $0.96~(6.45\cdot{10^{-24}})$ \\
\bottomrule
\end{tabular}
\end{adjustbox}
\end{sc}
\end{small}
\caption{Shift malignancy with statistical significance of domain classifier performance in parentheses. GRUD was used for both the domain classifier and determining the accuracy on the top 100 most anomalous samples for mortality and LOS. CNN was used for both the domain classifier and determining the accuracy on the the top 100 most anomalous samples for intervention prediction (Vaso). Malignant dataset shift appears in the earlier years in mortality and LOS. None of the shifts are malignant in intervention prediction.}
\label{tb2:nn}
\end{center}
\end{table}
\newpage
\section{Algorithm Definitions}
\label{appendix:algorithms}
\subsection{DP-SGD Algorithm}
\label{appendix:DP-SGD}
\begin{algorithm}[htb!]
   \caption{Differentially private SGD}
   \label{alg:diffprivacy}
\begin{algorithmic}
   \STATE {\bfseries Input:} Examples \{$x_1$,...,$x_N$\}, loss function $\mathcal{L}(\theta)$=$\frac{1}{N}\sum_{i=1}^{N} \mathcal{L}(\theta, x_i)$. Parameters: learning rate $\eta_{t}$, noise multiplier $\sigma$, mini-batch size $L$, $\ell_2$ norm bound $C$.
   \STATE \textbf{Initialize} $\theta_0$ randomly
   \FOR{$t$ $\in$ $[T]$}
   \STATE Take a random mini-batch $L_t$ with sampling probability $L/N$
   \STATE \textbf{Compute gradient}
   \STATE For each $i \in L_t$, compute $\mathbf{g_t}({x_i})\leftarrow\nabla_{\theta_t}\mathcal{L}(\theta, x_i)$ 
   \STATE \textbf{Clip gradient}
   \STATE $\bar{\mathbf{g_t}}({x_i})\leftarrow\mathbf{g_t}({x_i})/max\left(1,\frac{\left||\mathbf{g_t}({x_i})\right||}{C}\right)$ 
   \STATE \textbf{Add noise}
   \STATE $\tilde{\mathbf{g_t}}\leftarrow\frac{1}{L}\left({\sum_{i}\bar{\mathbf{g_t}}({x_i})}+\mathcal{N}\left(0,\sigma^2C^2\mathbf{I}\right)\right)$
   \STATE \textbf{Descent}
   \STATE $\theta_{t+1}\leftarrow {\theta_t} - \eta_{t}\tilde{\mathbf{g_t}}$
   \ENDFOR
   \STATE {\bfseries Output:} $\theta_T$ and the overall privacy cost $(\epsilon,\delta)$.
\end{algorithmic}
\end{algorithm}

\subsection{Objective Perturbation}
\label{appendix:obj_pert_alg}
\begin{algorithm}[htb!]
    \caption{Objective perturbation for differentially private LR}
    \label{alg:obj_pert}
   \begin{algorithmic}
      \STATE {\bfseries Inputs:} Data $\mathcal{D} = \{z_{i}\}$, parameters $\epsilon_p, \Lambda, \alpha, C$
      \STATE {\bfseries Output:} Approximate minimizer $\textbf{f}_{priv}$
      \STATE The weights of the linear model are defined as $\textbf{f}$
      \STATE The private loss function is defined as $J_{priv}(\textbf{f},\mathcal{D}) = J(\textbf{f},\mathcal{D}) + \frac{1}{n}\textbf{b}^{T}\textbf{f}$
      \STATE Normalize all the records in $\mathcal{D}$ by $C$
      \STATE Let $\epsilon_{p}^{'} = \epsilon_{p}$ - $log(1 + \frac{2c}{n\Lambda}$ + $\frac{c^2}{n^{2}\Lambda^{2}})$ 
      \STATE If $\epsilon_{p}^{'} > 0$, then $\Delta = 0$, else $\delta = \frac{c}{n(e^{{\epsilon_{p}/4}-1})} - \Lambda$, and $\epsilon_{p}^{'} = \epsilon_{p}/2$
      \STATE Draw a vector \textbf{b} according to $v$(\textbf{b}) = $\frac{1}{\alpha}e^{-\beta||\textbf{b}||}$ with $\beta = \epsilon_{p}^{'}/2$
      \STATE Compute $\textbf{f}_{priv}$ = argmin$J_{priv}(\textbf{f},\mathcal{D}) + \frac{1}{2}\Delta||\textbf{f}||^2$
   \end{algorithmic}
\end{algorithm}

\section{Training Setup and Details}
\label{appendix:training_setup}
We trained all of our models on a single NVIDIA P100 GPU, 32 CPUs and with 32GB of RAM. We perform each model run using five random seeds so that we are able to produce our results with mean and variances. Furthermore, each one is trained in the three privacy settings of none, low, and high. Each setting required a different set of hyperparameters which are discussed in the below sections. For all of the privacy models we fix the batch size to be 64 and the number of microbatches to be 16 resulting in four examples per microbatch.
\subsection{Logistic Regression}
LR models are linear classification models of low capacity
and moderate interpretability. Because LR does not naturally handle temporal data, 24
one-hour buckets of patient history are concatenated into one vector along with the static
demographic vector. For training with no privacy, we use the LR implementation in SciKit Learn’s LogisticRegression~\cite{pedregosa_scikit-learn:_2011}
class. We perform a random search over the following hyperparameters: regularisation
strength (C), regularisation type (L1 or L2), solvers (“liblinear” or “saga”), and maximum
number of iterations. For private training, we implement the model using Tensorflow~\cite{Abadi2016-lk} and Tensorflow Privacy. We use L2 regularisation and perform a grid search over the following hyperparameters: over the number of epochs (5 10 15 20) and learning rate (0.001 0.002 0.005 0.01). Finally, we use the DP-SGD optimizer implemented in Tensorflow Privacy to train our models. 

\subsection{GRU-D}
GRU-D models are a
recent variant of recurrent neural networks (RNNs) designed to specifically model irregularly
sampled timeseries by inducing learned exponential regressions to the mean for unobserved
values. Note that GRU-D is intentionally designed to account for irregularly sampled
timeseries (or equivalently timeseries with missingness). We implemented the model in Tensorflow based on a publicly available PyTorch implementation \cite{han-jd_2019}. For both not private and private training we use a hidden layer size of 67 units, batch normalisation, and dropout with a probability of 0.5 on the classification layer like in
the original work. We use the Adam optimizer for not private training and the DPAdam optimizer for private training with early stopping criteria for both.
\subsection{CNN}
We use three layer 1D CNN models with max pooling layers and ReLU activation functions. For both all models we perform a grid search over the following hyperparameters: dropout (0.1 0.2, 0.3, 0.4, 0.5), number of epochs (12, 15, 20), and learning rate (0.001, 0.002, 0.005, 0.01). Finally, we use the DPAdam optimizer for private learning.
\subsection{DenseNet-121}
We finetune a DenseNet-121 model that was pretrained on ImageNet using the Adam optimizer without DP and the DPAdam optimizer for private learning. We take the hyperparameters stated in~\cite{seyyed2020chexclusion}.

\section{Objective Perturbation Results}
\label{appendix:obj_pert}
\subsection{Utility Tradeoff}
\label{appendix:obj_pert_util}
\begin{table}[h]
\begin{center}
\begin{large}
\begin{sc}
\begin{adjustbox}{max width=\textwidth}
\begin{tabular}{llccc}
\toprule
\textbf{MIMIC-III} \\
\midrule
AUROC \\
\midrule
Task & Model & None ($\epsilon,\delta$) & Low ($\epsilon,\delta$) & High ($\epsilon,\delta$)\\
\midrule
Mortality & LR & $0.82 \pm 0.03~(\infty, 0)$ & $0.81 \pm 0.02~(3.50\cdot{10^5}, 0)$ & $0.54 \pm 0.03~(3.54, 0)$ \\
\midrule
Length of Stay > 3 & LR & $0.69 \pm 0.02~(\infty, 0)$ & $0.68 \pm 0.01~(3.50\cdot{10^5}, 0)$ & $0.57 \pm 0.03~(3.54, 0)$ \\
\midrule
AUPRC \\
\midrule
Mortality & LR & $0.35 \pm 0.05~(\infty, 0)$ & $0.31 \pm 0.06~(3.50\cdot{10^5},0)$ & $0.10 \pm 0.02~(3.54, 0)$ \\
\midrule
Length of Stay > 3 & LR & $0.66 \pm 0.03~(\infty, 0)$ & $0.65 \pm 0.03~(3.50\cdot{10^5}, 0)$ & $0.54 \pm 0.02~(3.54, 0)$ \\
\bottomrule
\end{tabular}
\end{adjustbox}
\end{sc}
\end{large}
\caption{privacy-utility tradeoff across vision and health care tasks. The health care tasks have a more significant tradeoff between the High and Low or None setting. The tradeoff is better in more balanced tasks (length of stay and intervention onset), and worst in tasks such as mortality.}
\label{tab:mimic_acc}
\end{center}
\end{table}
\subsection{Robustness Tradeoff}
\label{appendix:obj_pert_robust}
\begin{figure}[h]
    \centering
    \includegraphics[width=\textwidth]{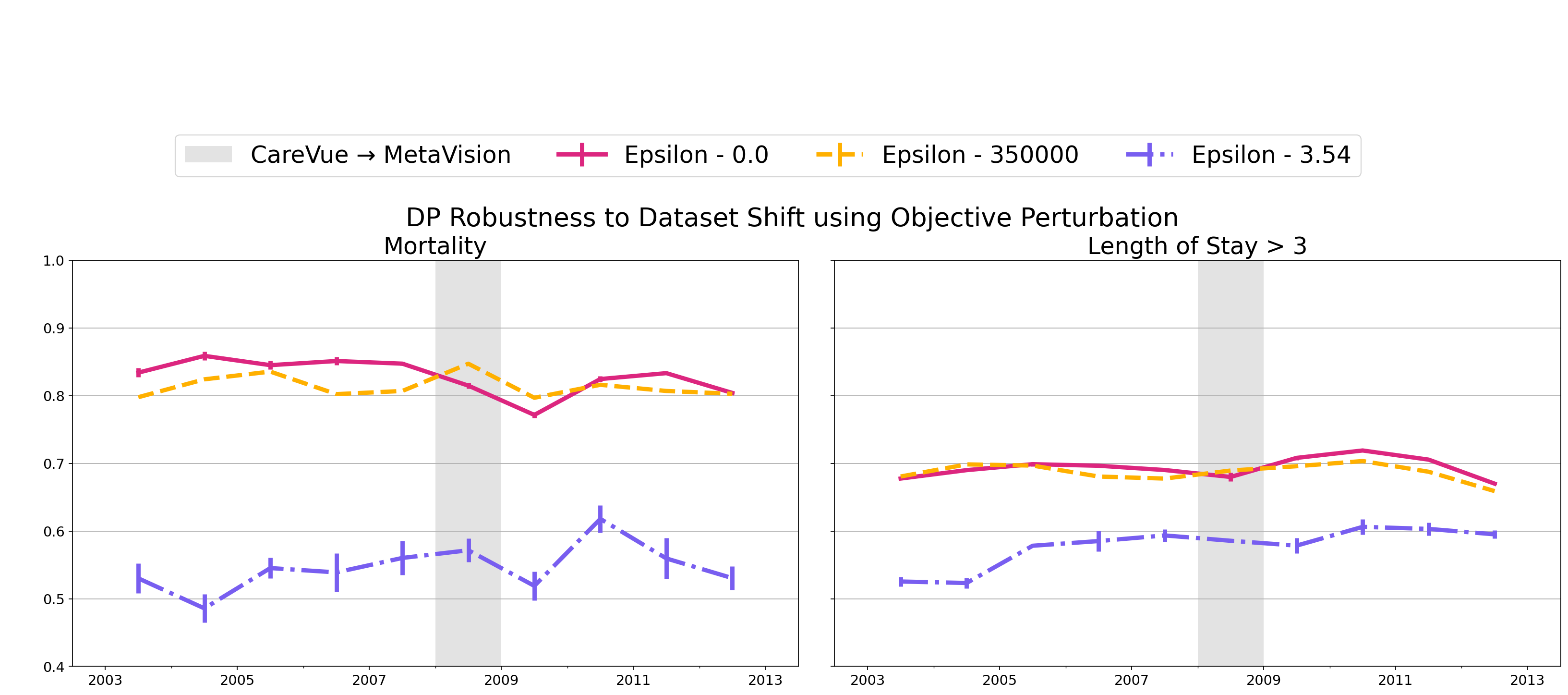}
    \caption{Characterizing the effect of DP learning on robustness to non-stationarity and concept shift. One instance of increased robustness in the 2009 column for mortality prediction in the high privacy setting (A), but this does not hold across all tasks and models. Performance drops in the 2009 column for LOS in both LR and GRUD (B), and a much worse drop in the high privacy CNN for intervention prediction (C).}
    \label{fig:mimic_priv_robust_lr_obj}
\end{figure}
\begin{figure}[h]
    \centering
    \includegraphics[width=\textwidth]{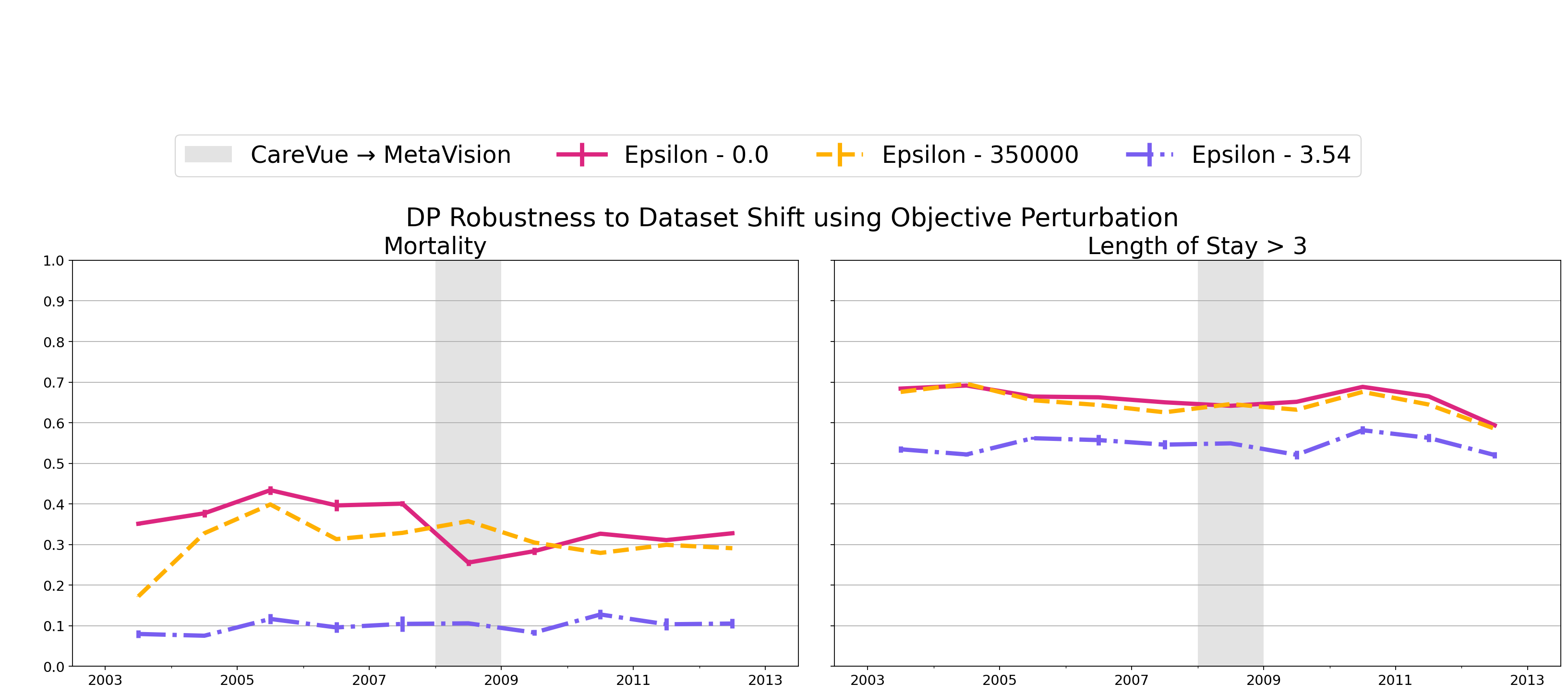}
    \caption{Characterizing the effect of DP learning on robustness to non-stationarity and concept shift. One instance of increased robustness in the 2009 column for mortality prediction in the high privacy setting (A), but this does not hold across all tasks and models. Performance drops in the 2009 column for LOS in both LR and GRUD (B), and a much worse drop in the high privacy CNN for intervention prediction (C).}
    \label{fig:mimic_priv_robust_lr_obj_apr}
\end{figure}

\newpage
\section{Additional DP-SGD Results}

\subsection{Health Care AUPRC Analysis}
\label{appendix:mimic_auprc}
\begin{figure}[h]
    \centering
    \includegraphics[width=\textwidth]{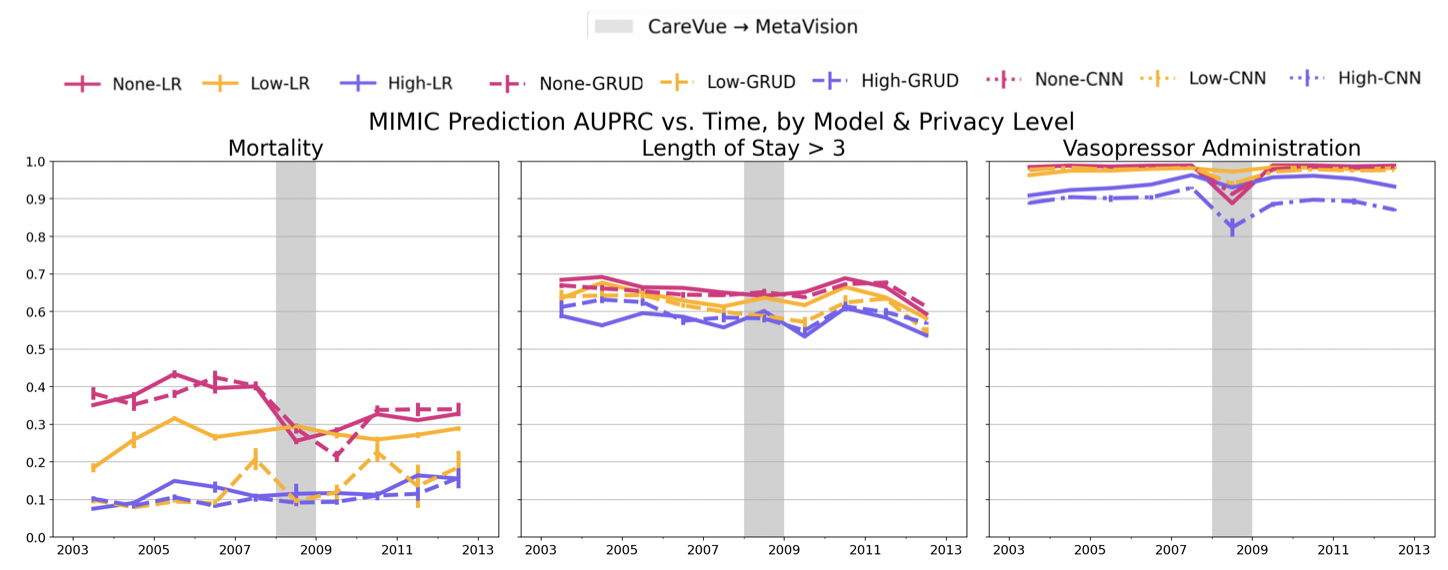}
    \caption{Characterizing the effect of DP learning on AUPRC robustness to non-stationarity and concept shift. One instance of increased robustness in the 2009 column for mortality prediction in the high privacy setting (A), but this does not hold across all tasks and models. Performance drops in the 2009 column for LOS in both LR and GRUD (B), and a much worse drop in the high privacy CNN for intervention prediction (C).}
    \label{fig:mimic_priv_robus_apr}
\end{figure}

In health care settings, the ability of the classifier to predict the positive class is important. We characterize the effect of differential privacy on this further by measuring the average performance across the years (Table~\ref{tab:mimic_apr}) and the robustness across the years (Fig.~\ref{fig:mimic_priv_robus_apr}) with area under the precision recall curve (AUPRC) and AUPRC (Micro) for the intervention prediction task.
\begin{table}[htb!]
\begin{center}
\begin{small}
\begin{sc}
\begin{adjustbox}{max width=\textwidth}
\begin{tabular}{llccc}
\toprule
Task & Model & None & Low & High\\
\midrule
Mortality & LR & $0.35 \pm 0.05$ & $0.26 \pm 0.05$ & $0.12 \pm 0.03$ \\
& GRUD & $0.35 \pm 0.06$ & $0.13 \pm 0.05$ & $0.11 \pm 0.02$ \\
\midrule
Length of Stay > 3 & LR & $0.66 \pm 0.03$ & $0.63 \pm 0.03$ & $0.57 \pm 0.03$ \\
& GRUD & $0.65 \pm 0.02$ & $0.61 \pm 0.03$ & $0.59 \pm 0.03$ \\
\midrule
Intervention Onset (Vaso) & LR & $0.98 \pm 0.03$ & $0.97 \pm 0.01$ & $0.93 \pm 0.03$ \\
& CNN & $0.98 \pm 0.02$ & $0.97 \pm 0.01$ & $0.89 \pm 0.03$ \\
\bottomrule
\end{tabular}
\end{adjustbox}
\end{sc}
\end{small}
\caption{privacy-utility tradeoff for AUPRC in health care tasks. The health care tasks have a more significant tradeoff between the High and Low or None setting. The tradeoff is better in more balanced tasks (length of stay and intervention onset), and worst in tasks such as mortality where class imbalance is present. There is a 23\% and 24\% drop in the AUPRC (Micro) between no privacy and high privacy settings for mortality prediction for LR and GRUD respectively.}
\label{tab:mimic_apr}
\end{center}
\end{table}
\label{sec:robustness}

\subsection{MIMIC-III Robustness Correlation Analysis}
\label{appendix:robust_corr}
To characterize a significant impact on robustness due to DP we perform a Pearson's correlation test between the generalization gap and the malignancy of shift. The generalization gap is measured as the difference between the classifier's performance on an in-distribution test set and an out of distribution test set. The results from this correlation analysis demonstrate that differential privacy provides no conclusive effect on the robustness to dataset shift (Table~\ref{tab:mimic_pearson}).
\begin{table}[htb!]
\begin{center}
\begin{small}
\begin{sc}
\begin{adjustbox}{max width=\textwidth}
\begin{tabular}{l|c|c|c|c}
\toprule
\textbf{Pearson's Correlation}  \\
\midrule
Task & Model & None & Low  & High\\
\midrule
Mortality & LR & $0.60~(0.05) $ & $0.51~(0.11)$ & $-0.03~(0.93)$ \\
& GRUD & $0.23~(0.50)$ & $0.55~(0.08)$ & $-0.24~(0.47)$\\
\midrule
Length of Stay > 3 & LR & $0.14~(0.68)$ & $0.30~(0.37)$ & $0.45~(0.16)$\\
& GRUD & $-0.57~(0.06)$ & $-0.51~(0.11)$ & $0.17~(0.61)$ \\

\midrule
Intervention Onset (Vaso) & LR & $0.27~(0.43)$ & $0.41~(0.21)$ & $-0.09~(0.79)$ \\
& CNN & $0.83~(0.00)$ & $0.57~(0.06)$ & $0.65~(0.03)$ \\
\bottomrule
\end{tabular}
\end{adjustbox}
\end{sc}
\end{small}
\caption{We calculate the pearson's correlation between AUROC gap and the malignancy of the shift. Positive correlations mean a lack of robustness since the generalization gap increases as the shift becomes more significant. Negative correlations mean improved robustness since the generalization gap decreases as the shift becomes more significant. We notice that the differential privacy improves robustness when a malignant shift is present in mortality but results in worse robustness in length of stay. None of the correlations are statistically significant so a claim cannot be made that differential privacy improves robustness to dataset shift in health care.}
\label{tab:mimic_pearson}
\end{center}
\end{table}
\newpage
\subsection{Fairness Analysis}
\subsubsection{Averages}
\label{appendix:mimic_fairness}
\begin{table*}[h]
\begin{center}
\begin{large}
\begin{sc}
\begin{adjustbox}{max width=\textwidth}
\begin{tabular}{lllccc}
\toprule
\textbf{AUROC Gap} & Protected Attribute & Model & None & Low & High \\
\midrule
Mortality & Ethnicity & LR & $0.04 \pm -0.01$ & $0.03 \pm 0.02$ & $-0.02 \pm 0.02$ \\
& & GRUD & $0.03 \pm 0.02$ & $-0.03 \pm 0.01$ & $-0.0072 \pm 0.03$ \\
\midrule
Length of Stay > 3 & Ethnicity & LR & $-0.003 \pm 0.002$ & $-0.009 \pm 0.005$ & $0.0007 \pm 0.003$  \\
& & GRUD & $-0.001 \pm 0.006$ & $0.006 \pm 0.005$ & $-0.001 \pm 0.004$ \\
\midrule
Intervention Onset (Vaso) & Ethnicity & LR & $-0.007 \pm 0.001$ & $-0.004 \pm 0.001$ & $0.002 \pm 0.005$  \\
& & CNN & $-0.001 \pm 0.002$ & $0.008 \pm 0.000$ & $0.000 \pm 0.002$ \\
\midrule
NIH Chest X-Ray & Sex & DenseNet-121 & $-0.014 \pm 0.021$ & $-0.001 \pm 0.006$ & $-0.0036 \pm 0.0086$ \\
\midrule
\textbf{Recall Gap}  \\
\midrule
Mortality & Ethnicity & LR & $-0.006 \pm 0.046$ & $0.000 \pm 0.002$ & $0.000 \pm 0.000$ \\
&  & GRUD & $0.013 \pm 0.067$ & $-0.043 \pm 0.046$ & $-0.002 \pm 0.059$ \\
\midrule
Length of Stay > 3 & Ethnicity & LR & $0.015 \pm 0.017$ & $0.024 \pm 0.018$ & $0.001 \pm 0.000$  \\
&  & GRUD & $0.054 \pm 0.039$ & $0.078 \pm 0.033$ & $0.053 \pm 0.056$ \\
\midrule
NIH Chest X-Ray & Sex & DenseNet-121 & $-0.000 \pm 0.005$ & $0.007 \pm 0.013$ & $0.001 \pm 0.013$ \\
\midrule
\textbf{Parity Gap}\\
\midrule
Mortality & Ethnicity & LR & $-0.046 \pm 0.018$ & $0.000 \pm 0.000$ & $0.000 \pm 0.000$ \\
& & GRUD & $0.013 \pm 0.009$ & $-0.007 \pm 0.012$ & $-0.021 \pm 0.022$ \\
\midrule
Length of Stay > 3 & Ethnicity & LR & $0.022 \pm 0.012$ & $0.037 \pm 0.015$ & $0.001 \pm 0.001$  \\
& & GRUD & $0.059 \pm 0.032$ & $0.071 \pm 0.017$ & $0.057 \pm 0.059$ \\
\midrule
NIH Chest X-Ray & Sex & DenseNet-121 & $0.001 \pm 0.007$ & $0.001 \pm 0.008$ & $0.002 \pm 0.006$ \\
\midrule
\textbf{Specificity Gap}\\
\midrule
Mortality & Ethnicity & LR & $0.058 \pm 0.019$ & $0.000 \pm 0.000$ & $0.000 \pm 0.000$ \\
& & GRUD & $-0.005 \pm 0.008$ & $0.007 \pm 0.011$ & $0.021 \pm 0.023$ \\
\midrule
Length of Stay > 3 & Ethnicity & LR & $-0.009 \pm 0.013$ & $-0.038 \pm 0.022$ & $-0.001 \pm 0.001$  \\
& & GRUD & $-0.046 \pm 0.032$ & $-0.082 \pm 0.023$ & $-0.052 \pm 0.064$ \\
\midrule
NIH Chest X-Ray & Sex & DenseNet-121 & $-0.001 \pm 0.008$ & $-0.001 \pm 0.007$ & $-0.001 \pm 0.006$ \\
\bottomrule
\end{tabular}
\end{adjustbox}
\end{sc}
\end{large}
\caption{The fairness gaps between white and Black patients across the different health care tasks, privacy levels and models. Positive values represent a bias towards the white patients and negative values represent a bias towards the Black patients. The models are more fair as the metric moves towards zero. The models are more unfair as the metric moves away from zero.}
\label{tbl:mimic_fairness}
\end{center}
\end{table*}
\newpage
\subsubsection{Time Series}
\begin{figure}[htb]
    \centering
    \includegraphics[width=0.7\textwidth]{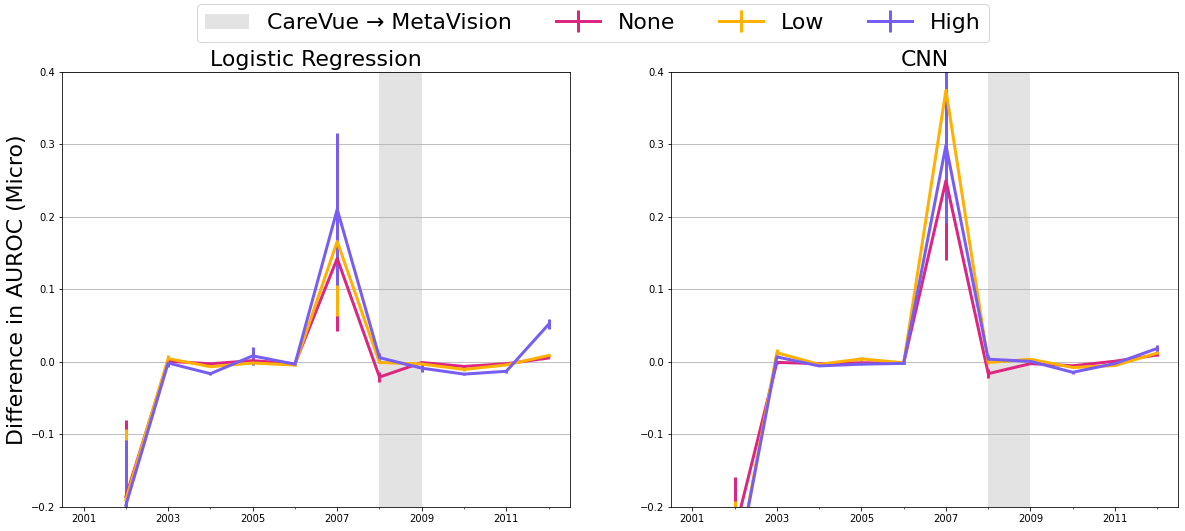}
    \caption{Characterizing the effect of differentially private learning on fairness with respect to non-stationarity and concept shift in the intervention prediction task. We find that all models experience similar performance with respect to the AUROC gap.}
    \label{fig:mimic_fair_time_intervention}
\end{figure}

\begin{figure}[htb]
    \centering
    \includegraphics[width=0.7\textwidth]{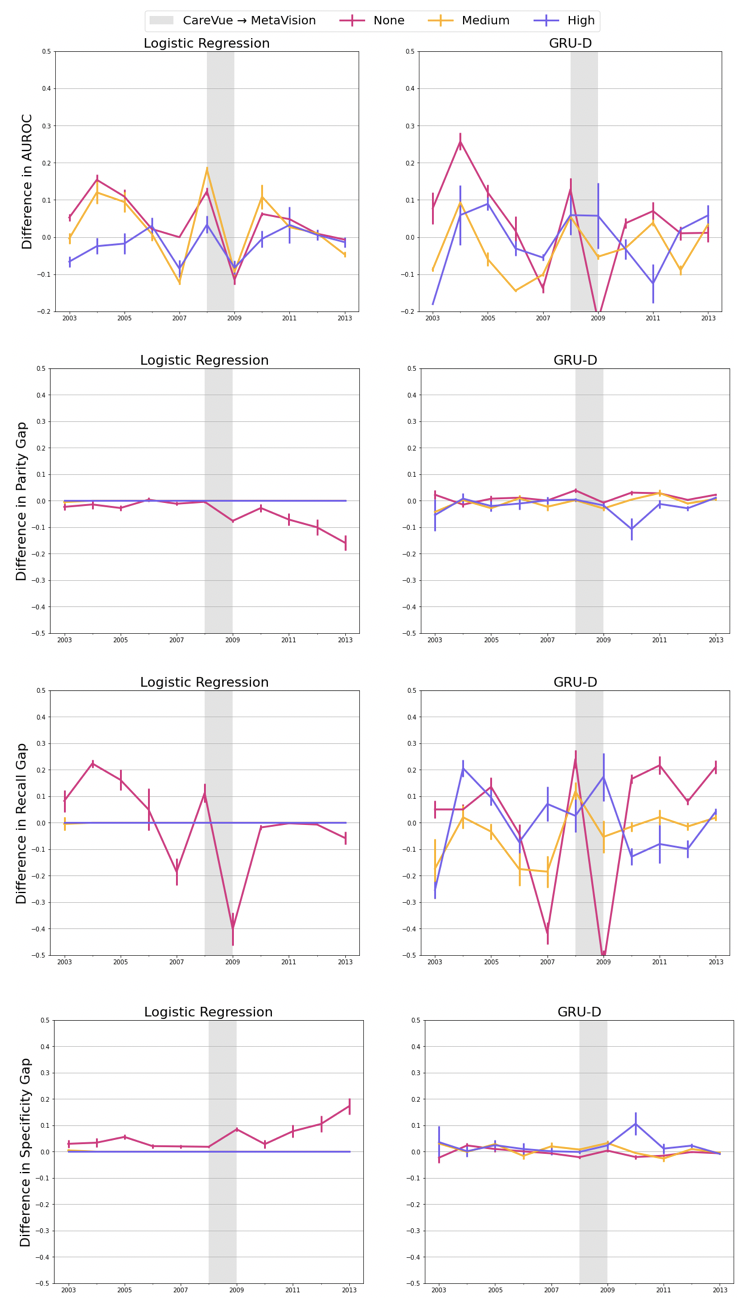}
    \caption{Characterizing the effect of differentially private learning on fairness with respect to non-stationarity and concept shift in the mortality prediction task. We find that the models trained without privacy experience high variance across the years across all definitions while models trained with privacy exhibit greater stability.}
    \label{fig:mimic_fair_time_mort}
\end{figure}

\begin{figure}[h]
    \centering
    \includegraphics[width=0.7\textwidth]{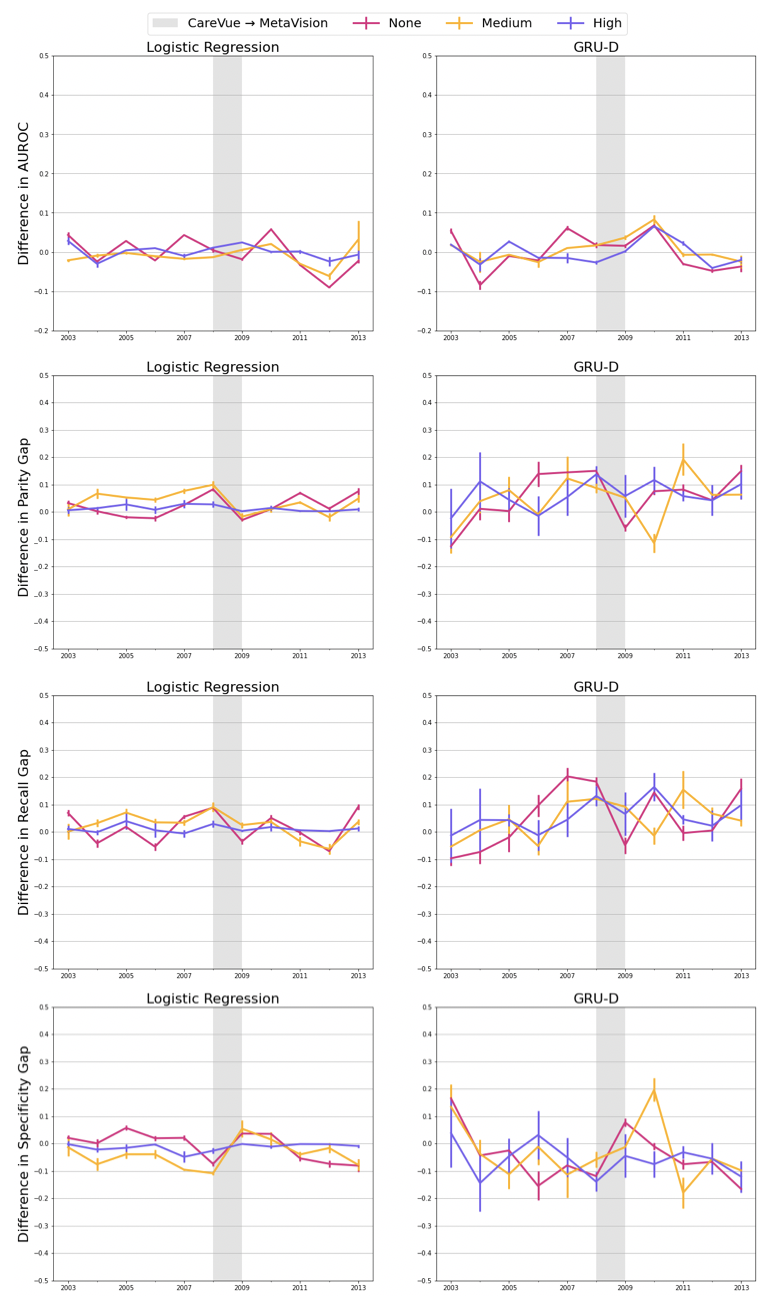}
    \caption{Characterizing the effect of differentially private learning on fairness with respect to non-stationarity and concept shift in the length of stay prediction task. We find that the models trained without privacy experience high variance across the years across all definitions while models trained with privacy exhibit greater stability.}
    \label{fig:mimic_fair_time_los}
\end{figure}

Although there is no effect on fairness from differential privacy when we average the metrics across all the years, we investigate how the metrics vary across each year we tested on. There is greater variance in the normal models than the private models for the parity, recall, and specificity gap (Fig~\ref{fig:mimic_fair_time_mort} and~\ref{fig:mimic_fair_time_los}). In intervention prediction, there is one spike in unfairness in 2007 which is seen across all levels of privacy (Fig.~\ref{fig:mimic_fair_time_intervention}).
\newpage
\section{Influence Functions}
\subsection{Fairness Graph}
\label{fig:fair_infl}
\begin{figure*}[htb!]
    \centering
    \includegraphics[width=\textwidth]{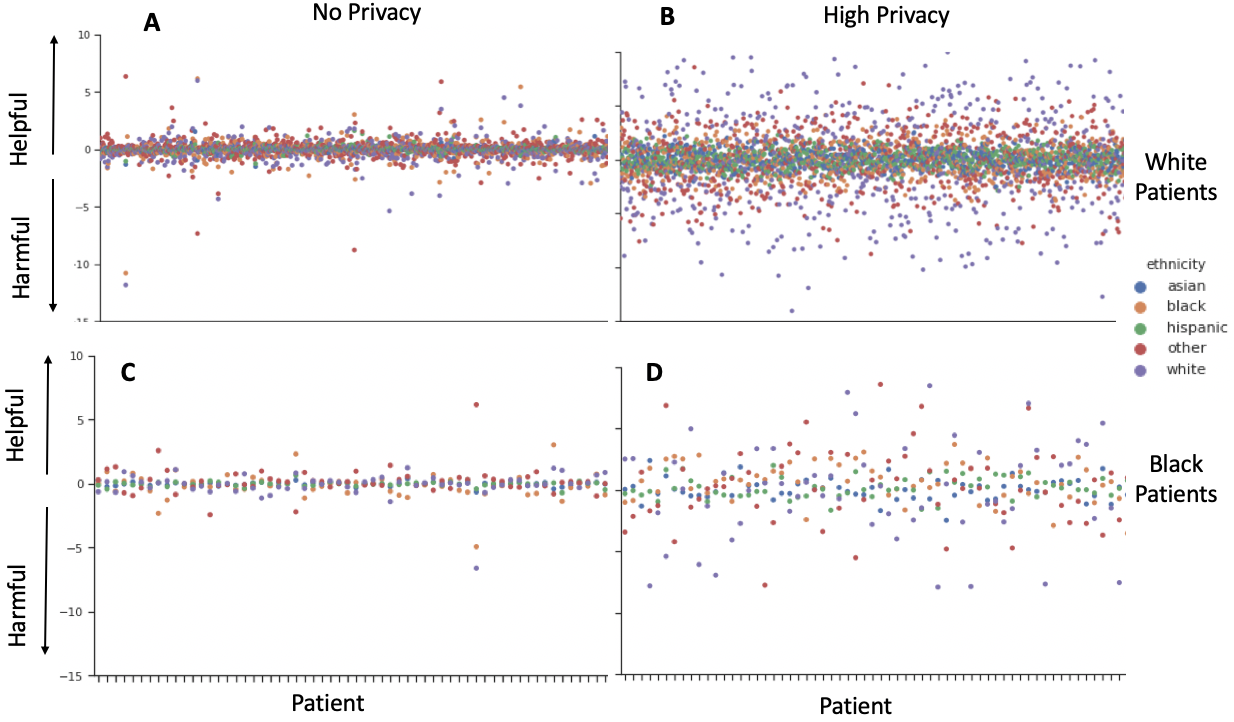}
  \caption{Group influence of training data per ethnic groups on 100 test patients with highest influence variance. The group influence of our majority ethnicity (white patients) is enhanced significantly in the high privacy setting, as demonstrated by the increased amplitude of those points in (B) and (D). In the no privacy setting the group influence of each ethnicity is similar for both white (A) and Black patients (C).}
    \label{fig:inf_fairness_graph}
\end{figure*}
\subsection{Privacy Makes the Most Harmful and Helpful Training Patients Personal}
We demonstrate that for the patients with the highest influence variance that their most helpful and harmful training patients are more common amongst other patients in the no privacy setting to them (Table~\ref{tab:most_helpful} and~\ref{tab:most_harmful}). This means that some patients carry their large influence around the test set of patients. This has important implications for differential privacy since it bounds the influence that anyone patient will have on the test loss of any other patient. The effect of the bounding is observed making the most helpful and harmful patients more personal to each patient in the high privacy setting (Table~\ref{tab:most_helpful_high} and~\ref{tab:most_harmful_high}).

\begin{table}[htb!]
\begin{sc}
\begin{adjustbox}{max width=\textwidth}
\centering
    \begin{tabular}{l|c}
    \toprule
    Most Helpful Patients (LR Mortality) \\
    \midrule
    Subject ID & Most Helpful Influence Count \\
    \midrule
    9980 & 24 \\
9924   &  13 \\
98995  &   8 \\
9954   &   5 \\
9905   &   5 \\
985    &   5 \\
990    &   4 \\
9929   &   4 \\
9998   &   4 \\
99726  &   3 \\
9942   &   3 \\
9896   &   2 \\
9873   &   2 \\
9867   &   2 \\ 
9825   &   2 \\
9937   &   2 \\
99938  &   2 \\
9893   &   2 \\
9932   &   1 \\ 
992    &   1 \\
99817  &   1 \\
98899  &   1 \\
9885   &   1 \\
98009  &  1 \\
977    &   1 \\ 
99485  &   1 \\
    \midrule
    \end{tabular}
\end{adjustbox}
\end{sc}
\caption{The frequency of the most helpful training patient in the first 100 patients with the highest influence variance for no privacy. Almost 25\% of the top 100 share the same most helpful training patient.}
\label{tab:most_helpful}
\end{table}

\begin{table}[htb!]
\begin{sc}
\begin{adjustbox}{max width=\textwidth}
\centering
    \begin{tabular}{l|c}
    \toprule
    Most Helpful Patients (LR Mortality) \\
    \midrule
    Subject ID & Most Helpful Influence Count \\
    \midrule
99938  &  7 \\
9994  &   5 \\
9977  &   4 \\
9970  &   4 \\
9998  &   3 \\
9991  &   3 \\
9987  &   3 \\
9973   &  3 \\
99528  &  3 \\
9965  &   3 \\
9949  &   2 \\
99598 &   2 \\
99469 &   2 \\
9889  &   2 \\
9980  &   2 \\
99817 &   2 \\
99384  &  2 \\
9984  &   2 \\
9937  &   2 \\
99883 &   2 \\
99726  &  2 \\
9950  &   2 \\
99936 &   2 \\
9983  &   2 \\
992  &    2 \\
9974  &   2 \\
9988  &   2 \\
9954  &   2 \\
9924  &   1 \\
9968  &   1 \\
9885 &    1 \\
9882  &   1 \\
9752 &    1 \\
9886  &   1 \\
99038 &   1 \\
99     &  1 \\
99063 &   1 \\
9951   &  1 \\
98698  &  1 \\
98919  &  1 \\
998   &   1 \\
9967  &   1 \\
9833  &   1 \\
9867   &  1 \\
9818  &   1 \\
9929   &  1 \\
99691 &   1 \\
9813   &  1 \\
9834  &   1 \\
9784   &  1 \\
9915   &  1 \\
9963   &  1 \\
9942   &  1 \\
9960   &  1 \\
    \midrule
    \end{tabular}
\end{adjustbox}
\end{sc}
\caption{The frequency of the most helpful training patient in the first 100 patients with the highest influence variance for high privacy. At most 7\% of the top 100 share the same most helpful training patient which is much less than the no privacy setting. Increasing privacy results in the most helpful patient being more personal to the test patient.}
\label{tab:most_helpful_high}
\end{table}

\begin{table}[htb!]
\begin{sc}
\begin{adjustbox}{max width=\textwidth}
\centering
    \begin{tabular}{l|c}
    \toprule
    Most Harmful Patients (LR Mortality) \\
    \midrule
    Subject ID & Most Hamrful Influence Count \\
    \midrule
    10013 &   22 \\
10015  &  19 \\ 
10007  &  13 \\
10045  &  10 \\
10036   &  8 \\
10076  &   5 \\
10088  &   4 \\
10077  &   4 \\ 
1004   &   3 \\
10102  &   3 \\
10028  &   2 \\
10038   &  2 \\
10173   &  2 \\
10027   &  1 \\
10184   &  1 \\
10289   &  1 \\
    \midrule
    \end{tabular}
\end{adjustbox}
\end{sc}
\caption{The frequency of the most harmful training patient in the first 100 patients with the highest influence variance for no privacy. 22\% of the top 100 share the same most harmful training patient.}
\label{tab:most_harmful}
\end{table}

\begin{table}[htb!]
\begin{sc}
\begin{adjustbox}{max width=\textwidth}
\centering
    \begin{tabular}{l|c}
    \toprule
    Most Harmful Patients (LR Mortality) \\
    \midrule
    Subject ID & Most Harmful Influence Count \\
    \midrule
   100    &   7 \\
10063  &   7 \\
10022  &   6 \\
10007  &   5 \\
1005   &   4 \\
10013  &   4 \\
10026  &   4 \\
10006   &  4 \\
10010  &   4 \\
10032   &  3 \\
10076   &  3 \\
10038   &  3 \\
10050   &  3 \\
10088   &  2 \\
10049   &  2 \\
10059   &  2 \\
10028   &  2 \\
10045   &  2 \\
10046   &  2 \\
10094  &   2 \\
1009   &   2 \\
10040   &  2 \\
10089   &  2 \\
10030   &  2 \\
10112   &  2 \\
10233  &   1 \\
10085   &  1 \\
101     &  1 \\
10114   &  1 \\
10149   &  1 \\
1028    &  1 \\
10320   &  1 \\
10110   &  1 \\
1020    &  1 \\
10335  &   1 \\
10043  &   1 \\
10044  &   1 \\
10266   &  1 \\
10042   &  1 \\
10015   &  1 \\
10122   &  1 \\
10217   &  1 \\
10165   &  1 \\
10123   &  1 \\
    \midrule
    \end{tabular}
\end{adjustbox}
\end{sc}
\caption{The frequency of the most harmful training patient in the first 100 patients with the highest influence variance for high privacy. At most 7\% of the top 100 share the same most harmful training patient which is much less than the no privacy setting. Increasing privacy results in the most harmful patient being more personal to the test patient.}
\label{tab:most_harmful_high}
\end{table}

\subsection{Years Analysis}
\label{appendix:years_inf_add}
We extend our analysis from the main paper by looking at the change in influence across each year for both no privacy and high privacy in the mortality task using LR.
\begin{figure}[htb!]
    \centering
    \includegraphics[width=\textwidth]{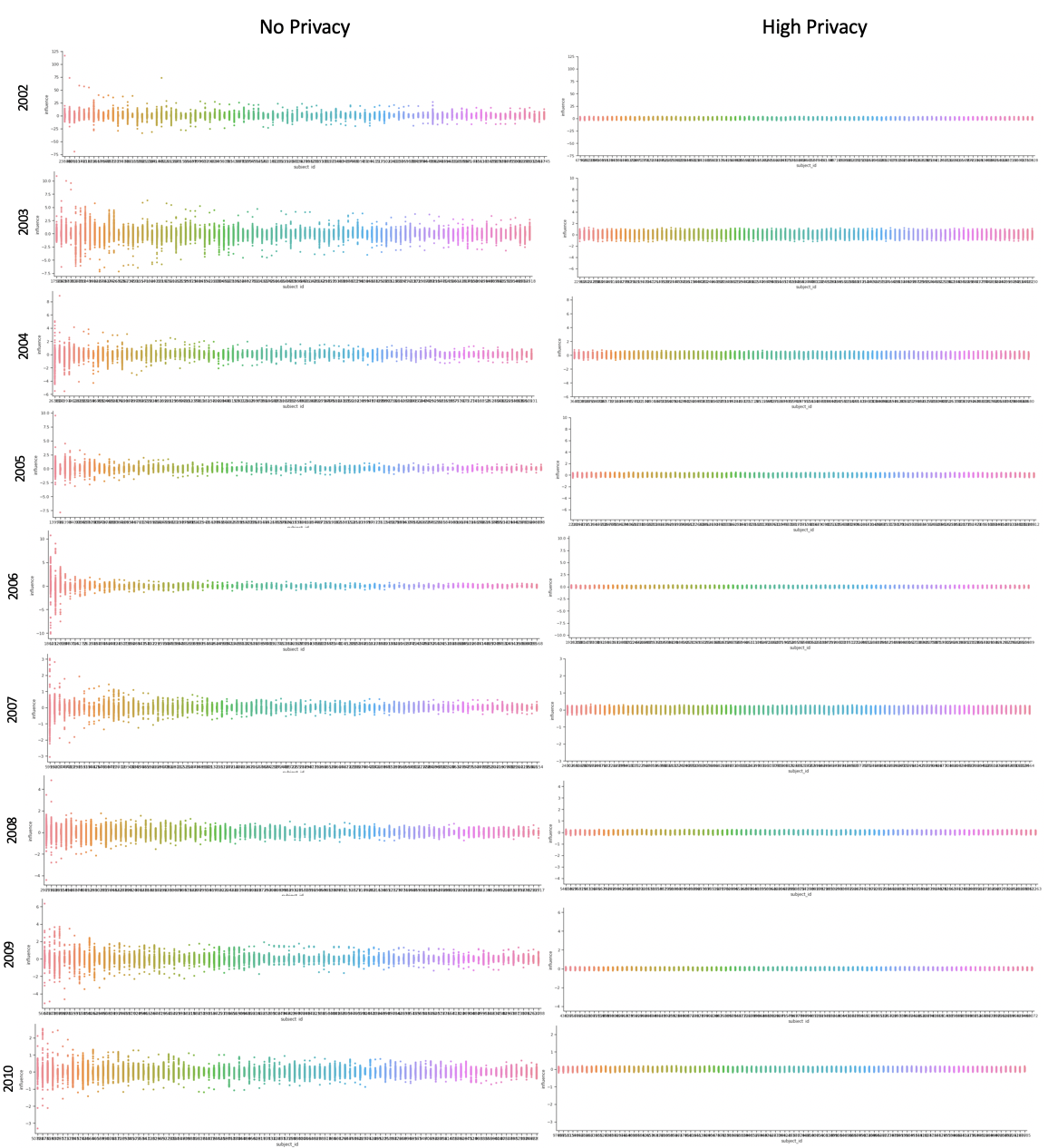}
  \caption{The progression of the influences of all training points on the test points across each year. The high privacy continues to bound the influence for each year agnostic to the dataset shift.}
    \label{fig:inf_fairness}
\end{figure}
\end{document}